\documentclass[11pt,draftclsnofoot,onecolumn]{IEEEtran}

\usepackage{amssymb,amsmath,color}
\usepackage{subfig,graphicx,colortbl,multicol,color}
\usepackage{eulervm}
\usepackage{tikz}
\usepackage{subfig}
\usepackage{cases}
\usepackage[numbers]{natbib}
\usepackage{setspace}

\usepackage{graphicx}
\usepackage{multirow}

\def\reg{{\rm\ooalign{\hfil
     \raise.07ex\hbox{\scriptsize R}\hfil\crcr\mathhexbox20D}}}
\DeclareMathOperator*{\argmin}{arg\,min}

\def\Rbb{\mathbb{R}}
\def\Cbb{\mathbb{C}}

\def\Scal{\mathcal{S}}
\def\figwidth{1\columnwidth}

\newcommand{\defeq}{\triangleq}

\usetikzlibrary{arrows}
\tikzstyle{c} = [circle,draw=black,minimum size=5ex]
\tikzstyle{r} = [rectangle,draw=black,minimum size=5ex]
\tikzstyle{a} = [->,>=stealth']

\def\reg{{\rm\ooalign{\hfil
     \raise.07ex\hbox{\scriptsize R}\hfil\crcr\mathhexbox20D}}}

\newlength \myfigwidth
\newlength \myfigwidthll
\newlength \myfigwidthllI

\newlength \descriptionwidth

\if@twocolumn
\else
\fi
\if@twocolumn
\else
\fi

\begin{document}

\title{Structured Sparsity Models for Multiparty Speech Recovery from Reverberant Recordings}

\author{Afsaneh~Asaei,~\IEEEmembership{Student~Member,~IEEE,}
	Mohammad~Golbabaee,~\IEEEmembership{Student~Member,~IEEE,\\}
	Herv\'e~Bourlard,~\IEEEmembership{Fellow,~IEEE}
        and~Volkan~Cevher,~\IEEEmembership{Senior Member,~IEEE}
        \thanks{Afsaneh Asaei and Herv\'e Bourlard are with Idiap Research Institute and also affiliated with \'Ecole Polytechnique F\'ed\'erale de Lausanne, Switzerland. Mohammad Golbabaee and Volkan Cevher are with \'Ecole Polytechnique F\'ed\'erale de Lausanne, Switzerland.\newline
	\indent E-mails: afsaneh.asaei@idiap.ch, mohammad.golbabaei@epfl.ch, herve.bourlard@idiap.ch, volkan.cevher@epfl.ch}\\
\thanks{Copyright (c) 2010 IEEE. Personal use of this material is permitted. However, permission to use this material for any other purposes must be obtained from the IEEE by sending a request to pubs-permissions@ieee.org.}
}

\maketitle

\begin{abstract}

We tackle the multi-party speech recovery problem through modeling the acoustic of the reverberant chambers. Our approach exploits structured sparsity models to perform room modeling and speech recovery. We propose a scheme for characterizing the room acoustic from the unknown competing speech sources relying on localization of the early images of the speakers by sparse approximation of the spatial spectra of the virtual sources in a free-space model. The images are then clustered exploiting the low-rank structure of the spectro-temporal components belonging to each source. This enables us to identify the early support of the room impulse response function and its unique map to the room geometry. To further tackle the ambiguity of the reflection ratios, we propose a novel formulation of the reverberation model and estimate the absorption coefficients through a convex optimization exploiting joint sparsity model formulated upon spatio-spectral sparsity of concurrent speech representation. The acoustic parameters are then incorporated for separating individual speech signals through either structured sparse recovery or inverse filtering the acoustic channels. The experiments conducted on real data recordings demonstrate the effectiveness of the proposed approach for multi-party speech recovery and recognition.

\end{abstract}


\IEEEpeerreviewmaketitle

\section{Introduction}
\label{sec:intro}
\IEEEPARstart{R}ECOVERY of speech signal from an acoustic clutter of unknown competing sound sources plays a key role in many applications involving distant-speech recognition, scene analysis, video-conferencing, hearing aids, surveillance, sound-field equalization and sound reproduction. Despite the vast efforts devoted to the issues arising in real-world conditions, development of systems to operate in the presence of competing sound sources yet remains a demanding challenge~\cite{SiSEC2011}. 

This paper considers distant-talking speech recognition in multi-party environment where multiple sound sources talk simultaneously. The common existence of overlapped speech segments has been shown to increase the speech recognition word error rate up to 30\% for a large vocabulary task~\cite{pas1} hence, it is required to incorporate an effective source separation technique to segregate the desired speech from the competing signals prior to recognition. We assume that the signals are acquired by an array of calibrated microphones. 

Previous approaches to multi-channel speech separation can be broadly dichotomized into three classes. The first category incorporates a prior knowledge about mutual independence and statistical characteristics of the source signals to identify the mixing model and to recover the individual sources~\cite{Ozerov-12, Douglas-05}. The method proposed in~\cite{TRINICON} exploit the statistical characteristics to estimate the acoustic channel of the enclosure and performs joint deconvolution and separation of speech signals. The underlying assumption of the approaches belonging to the first category is the statistical independence of the sources. Moreover, these techniques are confined to the scenarios where the number of microphones is greater than or equal to the number of sources also known as overdetermined or determined mixtures respectively~\cite{BSS}. 

The second category relies on spatial filtering techniques based on beamforming or steering a microphone array beam-pattern towards the target speaker thus resulting in the suppression of the undesired sources~\cite{mike_geom, Araki-07}. The underlying assumption of this approach is that there is no reverberation so the beamforming techniques are formulated upon upon direct path acquisition of the signals. These geometric techniques can work with any number of microphone including the scenarios in which the number of sources exceeds the number of sensors thereby, we have underdetermined mixtures~\cite{DSR-book}. 

The third category is based on sparse representation of the source signal, also known as Sparse Component Analysis (SCA)~\cite{spars1, Remi}. These techniques exploit a prior assumption that the sources have a sparse representation in a known basis or frame. The notion of sparsity opens a new road to address the underdetermined unmixing problem to estimate the unknown variables from a fewer number of known data. As there are many solutions to such systems, the answer ought to be the sparest solution measured in terms of the sparsity inducing norms~\cite{Remi, Zibulevsky, ellq}. The prior art on multichannel speech recovery through sparsity models are largely confined to the recovery of the signals at individual frequency level and ignore the higher-level structures exhibited in data representation.  

\begin{figure}[t]
  \centering
  \includegraphics[width=0.6\columnwidth]{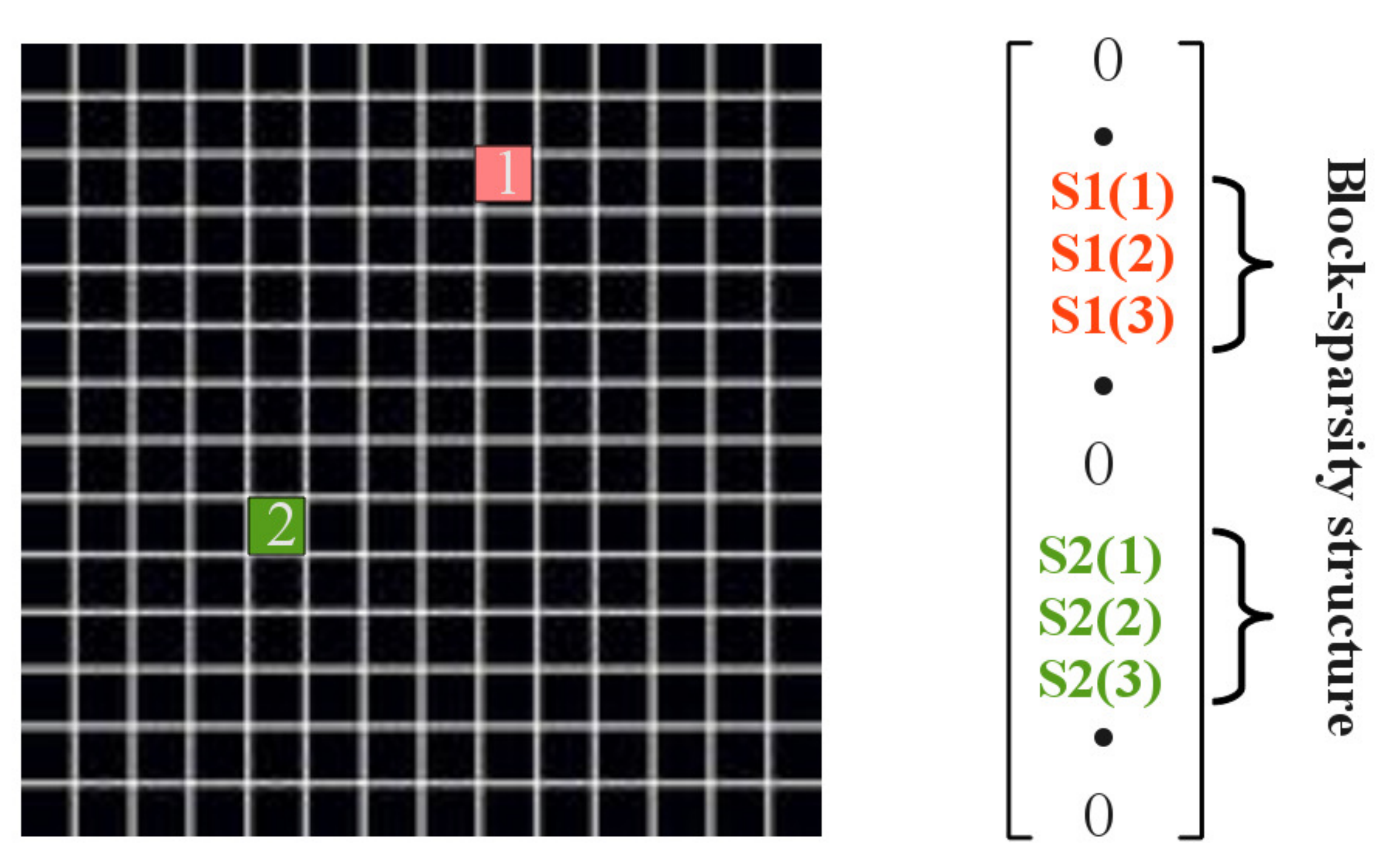}
  \caption{\small{The spatial sparsity of the speakers inside the room is illustrated through discretization of the planar area of the room into a grid of $G$ cells. The sources occupy only two cells marked as $1$ and $2$. Hence, the spatial representation of the source signals generated inside the room is sparse.\\ Assuming that the sources are immobile, if we denote the 3 instances of the signal attributed to the speaker at cell $i$ as $S_i(n) \in \Cbb^{3\times 1}, n \in \{1,2,3\}$ and concatenate the signals corresponding to each cell, the signal vector of the room can be formed as $\mathcal{S}=[S_1^T,...,S_G^T]^T \in \Cbb^{3G\times 1}$. We can see that support of $\mathcal{S}$ exhibits the block-sparsity structure as there are only two blocks of non-zero elements corresponding to the two speakers. The size of each block is the number of recording instances.}}
  \label{fig:grid}
\end{figure}

The approach that we propose in this paper relies on structured sparsity models underlying multiparty multi-channel recordings in reverberant environments. We discretize the planar area of the room into a grid of uniform cells where each of the speakers is located at one of the cells. If there are $N$ speakers in the room and given a fine gird of $G$ cells such that the cell's occupancy is exclusive, the distribution of the sources in the room is sparse; i.e., out of $G$ cells only \(N \ll G\) contain the sound sources. This implies the spatial sparsity model as depicted in Fig. \ref{fig:grid}.

Denoting the signal attributed to the source located at cell $i$ as $S_i$ and concatenating the signals corresponding to each cell, the signal vector coming from all over the room can be formed as $\mathcal{S}=[S_1^T,...,S_G^T]^T$ where $T$ stands for transpose. If we consider 1 instance of recordings from $N$ speakers, $\mathcal{S}$ is a sparse vector with only $N$ non-zero elements. The support of $\mathcal{S}$ corresponds to the $N$ cells where the sources are located. If we consider $F$ instances of recordings and assume that sources are immobile, each instance of the signal of a particular source implies sparsity in exactly the same manner as every other instances as they all correspond to the one particular cell where the source is located. This extra restriction imposes a constraint on the structure of the elements in $\mathcal{S}$ which goes beyond simple sparsity. We characterize sparsity with such constraints as structured sparsity. Fig \ref{fig:grid} illustrates the particular block sparsity model exhibited in representation of the signals coming from all over the grid as described here. 

This paper exploits structured sparsity models to recover the unknown individual speech signals: $S_i, i \in {1,...,G}$ from a few known multi-party recordings when the speakers are talking simultaneously. In addition to the spatial sparsity, we will exploit sparsity in spectral domain. The spectral structure of voiced speech typically comprises a small number of spectral peaks at harmonics of a fundamental frequency; at other frequencies the energy is typically low or negligible. We can therefore model the distribution of energy over frequencies as being sparse. Furthermore, we model the sparsity underlying the acoustic model of the room characterized by the \emph{Image Model} of multipath effect. The contribution of this paper is ultimately to introduce a unified theory of multiparty speech recovery formulated as a problem of signal recovery by exploiting structured sparsity models underlying the representation of information embedded in multichannel recordings. 

\section{State-of-the-Art}
\label{sec:art}
This paper tackles the multi-party speech recovery problem through modeling the acoustic of the enclosure and exploiting sparsity models. The room acoustic characterization was earlier incorporated in the method proposed in~\cite{TRINICON}. Their approach relies on statistical independence assumption of the sources to perform joint deconvolution and separation of speech signals and it is limited to over\/determined scenarios. This assumption has been relaxed in the method proposed in~\cite{Nesta-12} where multiple complex valued Independent Component Analysis (ICA) adaptations jointly estimate the mixing matrix and the temporal activities of multiple sources in each frequency band to exploit the spectral sparsity of speech signals. However, it does not explicitly rely on identification of the acoustic channel and recovery of the desired source imposes a permutation problem due to mis-alignment of the individual source components~\cite{Nesta-12}. 

A blind channel identification approach for speech separation and dereverberation is proposed in~\cite{Huang-Benesty-05}. In this paper, the mixing procedure is delineated with a multiple-input multiple-output (MIMO) mathematical model. The authors propose to decompose the convolutive source separation problem into sequential procedures to remove spatial interference at the first step followed by deconvolution of temporal echoes. To separate the speech interferences, the MIMO system of recorded overlapping speech in reverberant environment is converted into the single-input-multi-output (SIMO) systems corresponding to the channel associated with each speaker. The SIMO channel responses are then estimated using the blind channel identification through the unconstrained normalized multi-channel frequency-domain least mean square (UNMCFLMS) algorithm~\cite{Huang-Benesty-03} and de-reverberation can be performed based on the Bezout theorem (also known in the context of room acoustics as the multiple-input/output inverse-filtering theorem (MINT)~\cite{MINT}). A real-time implementation of this approach has been presented in~\cite{Rotili-10}, where the optimum inverse filtering is substituted by an iterative technique, which is computationally more efficient and allows the inversion of long RIRs in real-time applications~\cite{Rotili-10}. The major drawback of such implementation is that it can only perform channel identification from single talk periods and it requires a high input signal-to-noise ratio. 

Another approach to perform joint dereverberation and speech separation extends the maximum likelihood criteria applied in Weighted Prediction Error Method (WPE) for joint de-reverberation and separation of individual speech sources from determined and overdetermined mixtures~\cite{Yoshioka-10}. This method does not perform channel estimation and it does not perform well in estimation of the acoustic channel and assumes that source spectral components are uncorrelated across time frames.  It also relies on a single source assumption and thus can not achieve dereverberation when there are multiple sound sources~\cite{Nakatani-11}.\newline

This paper takes a new perspective to analysis of multi-channel recordings. We cast the microphone array acquisition as compressive sensing the information embedded in acoustic field and we leverage the theory of model-based sparse recovery for characterization of the acoustic measurements and recovering the speech components. Our approach features 4 contribution: 

\begin{itemize}
\item 
We separate the individual speech sources from the underdetermined convolutive mixtures exploiting various algorithmic approaches to structured sparse recovery while incorporating different types of spectral, spatial as well as acoustic multi-path structures.
\item 
We estimate the geometry of the reflective surfaces from recording of multiple unknown sources located at unknown positions exploiting sparse recovery and low-rank clustering techniques. 
\item 
We propose a new formulation of the reverberation model and estimate the absorption factors of the surfaces using structured sparse recovery. 
\item 
We analyze how the performance of speech recovery is entangled with the design of microphone array layout.
\end{itemize}

In this paper, we first overview the problem statement and characterization of multi-party multi-channel 
recordings in Section~\ref{sec:msr} along with the main assumption under consideration. The structured sparse speech recovery algorithms are described in Section~\ref{sec:msr_speech}. We set up the formulation of the structured sparse acoustic modeling in Section~\ref{sec:acoustic}.  
We elaborate on the theory of room geometry estimation in Section~\ref{sec:geom} and propose the approaches to absorption coefficient estimation in Sections \ref{sec:RIR} and \ref{sec:absorption}. The experimental analysis of the proposed techniques are discussed in Section~\ref{sec:tests}. The conclusions are drawn in Section~\ref{sec:final}.\newline

The notation used in this paper will be as follow:

\begin{itemize}
\item
$x_m$: signal of the $m^{th}$ microphone in time domain
\item
$X_m$: signal of the $m^{th}$ microphone in frequency domain
\item
$s_n$: signal of the $n^{th}$ source in time domain
\item
$S_n$: signal of the $n^{th}$ source in frequency domain
\item
$h_{mn}$: acoustic channel between the $m^{th}$ microphone and $n^{th}$ source in time domain 
\item
$H_{mn}$: acoustic channel between the $m^{th}$ microphone and $n^{th}$ source in frequency domain 
\item 
$\Phi$: microphone array manifold matrix; it characterizes the acoustic projections associated to the acquisition of source signals inside the enclosure 
\item
$l$: each time sample
\item
$f$: each frequency bin
\item
$F$: number of Fourier coefficients
\item
$\tau$: each frame of speech
\item
$\mathcal{T}$: number of speech frames 
\item
$\circledast$: convolution operation
\item
$T$: transpose operation
\item
$*$: conjugate transpose operation
\item
$\dag$: pseudo-inverse operation
\item
$N$: number of sources
\item
$M$: number of microphones
\item
$G$: number of cells
\item
$c$: speed of sound assumed to be constant
\item
$R$: order of reflections in a reverberant room
\item
$D$: number of reflective surfaces within the enclosure

\end{itemize}
\section{Multiparty Reverberant Recordings}
\label{sec:msr}

\subsection{Problem Statement}
\label{sec:problem}
In the present paper, we deal with the problem of separating the signals of an unknown number of speakers from multi-channel recordings in a reverberant room. 

We consider an approximate model of the acoustic observation as a linear convolutive mixing process, stated concisely as

\begin{equation}
 x_m(l) = \sum_{n=1}^{N}h_{mn} \circledast s_n(l), \quad m = 1,...,M
\label{fconv}
\end{equation}

This formulation is stated in time domain; to represent it in a sparse domain, we apply the Gabor expansion, i.e., the discrete Short-Time Fourier Transform (STFT)
of speech signals. Following from the convolution-multiplication property of the Fourier transform, the mixtures in frequency domain
can be written as

\begin{equation}
 X_m(f,\tau) = \sum_{n=1}^{N}H_{mn}S_n(f,\tau), \quad m = 1,...,M
\label{fconv_F}
\end{equation}

Our objective is to recover the individual source signals $S_.$ from the distant microphone recordings. There is no prior information about the number of sources and the acoustic mixing channels.  

\subsection{Multi-party Speech Representation}
\label{sec:sparsity}
We consider a scenario in which $N$ speakers are distributed in a planar area spatially discretized into a grid of $G$ cells. We assume to have a sufficiently dense grid so that each speaker is located at one of the cells thus $N \ll G$. The spatial spectra of the sources is defined as a vector with a sparse support indicating the components of the signal corresponding to each cell of the gird. 

We consider spectro-temporal representation of multi-party speech and entangle the spatial representation of the sources with the spectral representation of the speech signal to form vector $\mathcal{S} = [S_1^{T} ... S_G^{T}]^T \in \Cbb^{GF\times 1}$. Each $S_n \in \Cbb^{F\times 1}$ denotes the spectral representation or signal of the $n^{th}$ source (located at cell number $n$) in Fourier domain. We express the signal ensemble at microphone array as a single vector $\mathcal{X} = [X_1^{T} ... X_M^{T}]^{T}$ where each $X_m \in \Cbb^{F\times 1}$ denotes the spectral representation of recorded signal at microphone number $m$. The sparse vector $\mathcal{S}$ generates the microphone observations as $\mathcal{X} = \Phi \mathcal{S}$. $\Phi$ is the microphone array measurement matrix consisted of the acoustic projections associated to the acquisition of source signals located on the grid.  

\subsection{Acoustic Measurement Characterization}
\label{sec:free}
We assume the room to be a rectangular enclosure consisting of finite impedance walls. The point source-to-microphone impulse responses of the room are calculated using the \emph{Image Model} technique~\cite{ImageRev}. Taking into account the physics of the signal propagation and multi-path effects, the projections associated with the source located at the cell $g$ where $\nu_g$ represents the position of the center of the cell and captured by microphone $i$ located at position $\mu_i$ are characterized by the media Green's function and denoted as $\xi_{\nu_g\rightarrow\mu_i}$ defined by

\begin{equation}
\begin{split}
 & \xi_{\nu_g\rightarrow\mu_i}^{f}:\\
 & X(f,\tau) = \sum_{r=1}^{R}\frac{\iota^r}{\|\mu_i-\nu_g^r\|^{\alpha}}\exp(-jf\frac{\|\mu_i-\nu_g^r\|}{c})S(f,\tau),
\end{split}
\label{multipath1}
\end{equation}
where $j = \sqrt{-1}$ and $\iota$ is the reflection coefficient; $\iota^r$ is the reflection coefficient after $r$ reflections of the walls. The attenuation constant $\alpha$ depends on the nature of the propagation and is considered in our model to equal 1 which corresponds to the spherical propagation. This formulation assumes that if $s_1(l)=s(l)$ and $s_2(l)=s(l-\varrho)$, then $S_2(f,\tau) \approx \exp(-jf\varrho)S_1(f,\tau)$.

Given the source-sensor projection defined in Equation (\ref{multipath1}), we construct matrix $\Xi_{\nu_g\rightarrow\mu_i}$ for the measurement of the $F$ consecutive frequencies as  

\begin{equation}\label{dici_Xi}
\Xi_{\nu_g\rightarrow\mu_i} =
\begin{bmatrix}
\xi_{\nu_g\rightarrow\mu_i}^{1} &  0  & \ldots & 0\\
0  &  \xi_{\nu_g\rightarrow\mu_i}^{2} & \ldots & 0\\
\vdots & \vdots & \ddots & \vdots\\
0  &   0       &\ldots & \xi_{\nu_g\rightarrow\mu_i}^{F}
\end{bmatrix}
\end{equation}

Hence, the projections associated to the acquisition of the source signals located on the grid by microphone $i$ is $\phi_i = [\Xi_{\nu_1\rightarrow\mu_i} ... \Xi_{\nu_g\rightarrow\mu_i} ... \Xi_{\nu_G\rightarrow\mu_i}]$ and the measurement matrix of $M$-channel microphone array would be defined as 

\begin{equation}\label{dici}
\Phi =
\begin{bmatrix}
\phi_1\\
\\
\vdots\\
\phi_M
\end{bmatrix}
\end{equation}

As indicated by Equation \ref{multipath1}, characterizing the acoustic projections amounts to identifying the location of the \emph{source images} as well as the absorption factors of the reflective surfaces. We exploit this parametric model to address the speech recovery problem in this paper. 

In Section \ref{sec:sparsity} we mentioned that the sparse vector $\mathcal{S}$ generates the microphone measurements as $\mathcal{X} = \Phi \mathcal{S}$. Our goal is to recover $\mathcal{S}$ from a small number of measurements (i.e., $M < G$). There are many solutions to this problem, we thus exploit the prior information on sparse properties of $\mathcal{S}$ to circumvent the ill-posedness of the problem. 

We cast the underdetermined speech recovery problem as sparse approximation where we exploit the underlying structure of the sparse coefficients to recover the signal components more efficiently from fewer number of measurements~\cite{MCS}. This is the topic of the following Section~\ref{sec:msr_speech}.

\section{Structured Sparse Speech Recovery}
\label{sec:msr_speech}
The goal is to estimate the structured sparse coefficient vector $\mathcal{S}$ such that $\mathcal{X} = \Phi \mathcal{S}$. This problem could be stated precisely as 

\begin{equation}
 \hat{\mathcal{S}} = \underset{\mathcal{S} \in \mathbb{M}}{\operatorname{argmin}} \|\mathcal{S}\|_0 \quad s.t. \quad \mathcal{X} = \Phi \mathcal{S} 
\label{msr_0}
\end{equation}
where $\mathbb{M}$ specifies the union of all vectors with a particular support structure. The counting function $\|.\|_0: \Rbb^G \rightarrow \Rbb$ returns the number of non-zero components in its argument. 

The major classes of computational techniques for solving sparse approximation problem stated in Equation (\ref{msr_0}) include greedy pursuit, convex relaxation, non-convex optimization, and Bayesian algorithms~\cite{TroppWright}. This paper considers greedy algorithms and convex optimization, which offer provable correct solutions under well-defined conditions. The greedy pursuit method iteratively refines the current estimate for the coefficient vector $\mathcal{S}$ by modifying one or several coefficients chosen to yield a substantial improvement in quality of the estimated signal. The Convex optimization approach solves a convex relaxation of Equation (\ref{msr_0}) by replacing the counting function with a sparsity inducing norm. 

\subsection{Structured Sparsity models}
We focus on two types of structures underlying the sparse coefficients: \newline

\begin{itemize}

\item 
The first structure is the block-dependency model which is exhibited if some interconnections between the adjacent frequencies exist. In case of the vector $\mathcal{S}$, the \emph{block sparsity structure} indicates that the spatial sparsity structure is the same at all neighboring discrete frequencies. In other words, a block of $b$ consecutive frequencies corresponds to the same cell so the signal of the individual sources is recovered with a structure of independent blocks defined as

\begin{equation}
 \mathcal{F}_{B} = \{ [f_1,...,f_b], [f_{b+1},...,f_{2b}], [f_{F-b+1},...,f_{F}]\}.
\label{block_definition}
\end{equation} \newline

\item
The second structure is the harmonic-dependency model exhibited if there are some interconnections between frequencies which are the harmonics of a fundamental frequency. In voiced speech, most of the energy in the speech signal occurs at harmonics of a fundamental frequency. The \emph{harmonic sparsity structure} captures this model: it indicates that at any cell of the grid, energy is present in all frequencies that can be expressed as harmonics of a fundamental frequency. To state it more precisely, the support of vector $\mathcal{S}$ has the following $\mathcal{F}_H$ structure defined as 

\begin{equation}
 \mathcal{F}_{H} = \{ k f_0 | 1 < k < K \},
\label{har_definition}
\end{equation} 
where $f_0$ is the fundamental frequency and $K$ is the number of harmonics. 

\end{itemize}

\subsection{Model-based Sparse Recovery}
The model-based sparse recovery algorithms have been proposed to incorporate the underlying structure of the sparse coefficients in recovering the unknown sparse vector. We use the model-based sparse recovery algorithms explained as follow: 

\begin{itemize}

\item
\textit{IHT}: Iterative hard thresholding (IHT) offers a simple yet effective approach to estimate the sparse vectors \cite{ThomasMike}. It seeks an $N$-sparse representation $\hat{\mathcal{S}}$ of the observation $\mathcal{X}$ iteratively to minimize the residual error. We use the algorithm proposed in~\cite{V_CAMSAP} which is an accelerated scheme for hard thresholding methods with the following recursion

\begin{equation}
\left\{
\begin{array}{l}      
   \hat{\mathcal{S}}_0 = 0 \\
   r_i = \mathcal{X} - \Phi \hat{\mathcal{S}}_i \\
   \hat{\mathcal{S}}_{i+1} = \mathcal{M}^{\mathcal{F}_.} \left(\hat{\mathcal{S}}_i + \kappa \Phi^T r_i \right)
\end{array}\right.
\end{equation}
where the step-size $\kappa$ is the Lipschitz gradient constant to guarantee the fastest convergence speed \cite{Nestrov10}. To incorporate for the underlying structure of the sparse coefficients, the model approximation $\mathcal{M}^{\mathcal{F}_.}$ is defined as reweighting and thresholding the energy of the components of  $\hat{\mathcal{S}}$ with $\mathcal{F}_B$ or $\mathcal{F}_H$ structures~\cite{V_CAMSAP}. \newline 

\item
\textit{OMP}: The Orthogonal Matching Pursuit (OMP) is a greedy pursuit algorithm which iteratively refines a sparse solution by successively identifying one or more components that yield the greatest improvement in quality. To describe our model-based OMP in mathematical formulation, we consider an index set $\varLambda$ which selects a subset of columns from $\Phi$. Denoting the set difference operator as $\backslash$, the columns of $\Phi_{\backslash \varLambda}$ corresponding to either $\mathcal{F}_B$ or $\mathcal{F}_H$ structures are searched per iteration and $\varLambda$ is expanded so as the mean-squared error of the signal approximation is minimized~\cite{Trop07,ThomasMike,HMP03}. The signal estimation algorithm would thus have the following recursion 

\begin{equation}
\left\{
\begin{array}{l}      
   \varLambda_0^{\mathcal{F}_.} = 0 \\
   \lambda_i = \underset{\lambda \in \Phi_{\backslash \varLambda_{i-1}^{\mathcal{F}_.}}}{\operatorname{argmin}} \|\mathcal{X} - \Phi_{\varLambda_{i-1}^{\mathcal{F}_.} \cup \lambda} \Phi_{\varLambda_{i-1}^{\mathcal{F}_.} \cup \lambda}^\dag \mathcal{X} \|_2 \\
   \varLambda_i^{\mathcal{F}_.} = \varLambda_{i-1}^{\mathcal{F}_.} \cup \lambda_i \\
   \hat{\mathcal{S}_i} = \Phi^\dag_{\varLambda_i} \mathcal{X}
\end{array}\right. 
\end{equation}

\item
\textit{$L_1L_2$}: Another fundamental approach to sparse approximation replaces the combinatorial counting function in the mathematical formulation stated in Equation (\ref{msr_0}) with the $L_1$ norm, yielding convex optimization problems that admit a tractable algorithm referred to as basis pursuit \cite{spg_siam}. We use a multiple-measurement version of basis pursuit algorithm by re-arranging the components of $\mathcal{S}$ as a row-sparse matrix with the $n^{\mathcal{F}_.}$ columns corresponding to the common sparsity structure $\mathcal{F}_.$ referring to either $\mathcal{F}_B$ or $\mathcal{F}_H$. Hence, the optimization problem to recover the structured sparse coefficients would be the following

\begin{equation}\label{spg_group}
\begin{split}
 \hat{\mathcal{S}}& = \argmin \|\mathcal{S}\|_{L_1,L_2} \quad  s.t. \quad \mathcal{X} = \Phi \mathcal{S},\\
  & \|\mathcal{S}\|_{L_1,L_2} = \left(\sum_{i=1}^{G}\left[\sum_{j=1}^{n^{\mathcal{F}_.}}\mathcal{S}^{2}(i,j)\right]^{1/2}\right).
\end{split}
\end{equation}

\end{itemize}

The speech recovery approach as described in this section, requires identification of the acoustic measurements. To tackle this problem, we incorporate the \textit{Image Model} of multipath effect, as stated in Equation (\ref{multipath1}). We elaborate on characterization of the room acoustic in the next section. 

\section{Structured Sparse Acoustic Modeling}
\label{sec:acoustic}
Recall from Section \ref{sec:free} that characterizing the acoustic projections amounts to identifying the location of the \emph{source images} as well as the    \emph{absorption factors} of the reflective surfaces. In Section \ref{sec:geom}, we estimate the geometry of the room to identify the location of the \emph{source images}. In Sections \ref{sec:RIR} and \ref{sec:absorption}, we address the problem of \emph{absorption coefficient} estimation.  

\subsection{Estimation of the Room Geometry}
\label{sec:geom}
The projection expressed in Equation (\ref{multipath1}) corresponds to characterization of the forward model of the room acoustic channel as

\vspace{-1mm}
\begin{equation}
 H(f,\mu_i,\nu_g) = \sum_{r=1}^{R}\frac{\iota^r}{\|\mu_i-\nu_g^r\|}\exp(jf\frac{\|\mu_i-\nu_g^r\|}{c}).
\label{multipath2}
\end{equation}
$H(f,\mu_i,\nu_g)$ indicates the room impulse response function between the microphone located at $\mu_i$ and a source located at $\nu_g$. Hence, identifying the locations of the $R$ Images of the source corresponds to identifying the support of the room impulse response function. According to the \emph{Image Model}, if the geometry of the enclosure is known, it is possible to identify the \emph{source images} up to any arbitrary order~\cite{ImageRev}. 

Recent studies have shown that the impulse response function is a unique signature of the room and the geometry can be reconstructed given that up to second order of reflections are known~\cite{hearRoom}. Relying on this observation, we propose to localize the \emph{source images} using the sparse recovery algorithm with a free space measurement model, i.e., $R=0$, while the deployment of the grid captures the location of early reflections. The support of the acoustic channel, $\{ \nu_r | 1<r<R \}$ corresponds to the cells where the recovered energy of the signal is maximized. We consider the localized source signals in a \emph{close proximity} to the microphone array within a distance $d$ as the actual sources generating the signals $S_n, n=1,...,I$. The localized images are sorted up to the order of $D(D+1)/2$ according to the Cosine angle between the estimated signals and the source signal ($S_n$) and considered as the images associated to the $n^{th}$ source. Given the 
location of the source images, we estimate the room geometry by brute-force search to identify the dimensions which generate the least-squares approximation of the location of source images from the location of the actual sources. Table \ref{tab:geometry} summarizes the steps to implement room geometry estimation.

\begin{table}[t]\small{
\caption{Room geometry estimation procedure}
\noindent \hrule height 0.9pt
\vspace{1.5mm}
\begin{itemize}

\item Run sparse source localization algorithm with a free-space measurement model.
\item Run k-means clustering using Cosine angle as the distance metric.
\begin{itemize}
  \item The centroid of the clusters are selected as the location of the nearest sources to the center of the array.
  \item The Cosine angle is measured between components of the signal attributed to the actual sources w.r.t. the components corresponding to the source images.
  \item The number of each cluster members is limited to $D(D+1)/2$.
\end{itemize}
\item Find the room geometry by identifying the dimensions which yield the best approximation of the location of source images in least-squares sense.

\end{itemize}
\vspace{1.5mm}
\noindent \hrule height 0.9pt
\label{tab:geometry}}
\end{table}

The approach that we presented in this section can estimate the room geometry if a single source or multiple unknown sources exist in the room. The \emph{Image Model} indicates the sparsity of the room impulse response function with a particular structure imposed by the vertical reflections. We refer to this property as the acoustic structured sparsity and exploit it to address the problem of estimating the absorption coefficients. In Section \ref{sec:RIR}, we propose an approach for estimating the absorption factors if there is only a single talker. In Section \ref{sec:absorption}, we elaborate on a novel model of the room reverberation which enables accurate estimation of the absorption factors from a plurality of speech sources.

\subsection{Single-Source Absorption Coefficient Estimation}
\label{sec:RIR}
We consider the linear convolutive model of the reverberant enclosure and denote the time-domain acoustic channel between the source and microphone $i$ as $h_i(l)$. Hence, the signal of the microphone $i$, $x_i(l)$ is a filtered version of source signal $s(l)$ as $h_i(l) \circledast s(l)$. It is straightforward to see that 

\begin{equation}\label{cr2}
\small
 x_i(l) \circledast h_j(l) = x_j(l) \circledast h_i(l);
\end{equation}

Considering an $L$-tap acoustic filter, for $l = L,...,\mathcal{L},$ where $\mathcal{L}$ is the length of the recorded signal, (\ref{cr2}) becomes:

\begin{equation}\label{cr3}
\small
 [\chi_i(L) -\chi_j(L)] \begin{bmatrix} h_j \\ h_i\end{bmatrix} = 0,
\end{equation}
where $h := [h(L),...,h(0)]^T$ and 

\begin{equation}\label{cr4}
\small
 \chi(L) = \begin{bmatrix}
x(L) & x(L+1) & \ldots & x(2L)\\
x(L+1) &  x(L+2) & \ldots & x(2L+1)\\
\vdots & \vdots & \ddots & \vdots\\
x(\mathcal{L}-L) & x(\mathcal{L}-L+1) &\ldots & x(\mathcal{L})
\end{bmatrix}.
\end{equation}

This equation forms the basic idea for blind channel identification by least squares optimization~\cite{CR}. Relying on the structured sparsity model as indicated by the \emph{Image Model} of multipath effect, we propose the optimization algorithm constrained on the structured sparsity to capture the main reflections characterized by the \emph{Image Model}. 

Despite the existence of various reflective objects inside the room, the structured sparsity model obtained through the room geometry is theoretically sound due to the fact that the multi-path signal energy is a function of the reflective areas. Hence, for the general environment of the meeting rooms, many objects are acoustically transparent~\cite{L1_room}. In addition to the theoretical evidence, we empirically verified the effectiveness of the structured sparsity constraint for identification of the real acoustic impulse responses from noisy reverberant data generated by the impulse responses available in Aachen Impulse Response (AIR) database~\cite{AIR}.

Given the room geometry and the source location, the support of the highest energy components of RIR is determined by the \emph{Image Model} and denoted by $\Omega_d$ which refers to the direct path component calculated precisely as $\vartheta$ and $\Omega_\mathcal{I}$ which refers to the support of the reflections. We define $\Pi := [\chi_i(L)\quad -\chi_j(L)]$ and $\mathcal{H} := [ h_j^T\quad h_i^T]^T$. The structured sparse acoustic filter will be obtained by the following optimization

\begin{equation}\label{cr5}
\begin{split}
&\hat{\mathcal{H}} = \argmin \| \mathcal{H} \|_1 \quad\\
 & s.t. \quad \| \Pi \mathcal{H} \|_2 \leq \epsilon, \quad \mathcal{H}(\Omega_d)=\vartheta, \quad \mathcal{H}(\Omega_r)>0 
\end{split}
\end{equation} 

The estimated RIR is then used to estimate the absorption coefficients of our model stated in Equations (\ref{multipath1}-\ref{dici}) by least squares fitting and to characterize the acoustic channels of all cell positions in order to identify the microphone array measurement matrix. The speech recovery is then achieved by structured sparse recovery algorithms as explained in Section \ref{sec:msr_speech}. 

\subsection{Multi-Source Absorption Coefficient Estimation}
\label{sec:absorption}
This section elaborates on a novel formulation of the reverberant recordings which entangles structured sparsity indicated by the \emph{Image Model} and the spatio-spectral sparsity of multiparty recordings for joint estimation of the absorption coefficients and recovery of the sources. Generalized to the algorithm proposed in Section \ref{sec:RIR}, we can estimate the frequency-dependent absorption factors in a multi-source environment. \newline

\subsubsection{Factorized Formulation of the Reverberant Recordings}
\label{sec:factor}
We propose a novel formulation of the \emph{reverberation model} factorized into permutation (corresponding to the \emph{source images}) and attenuation (corresponding to the absorption factors) of the sources in an unbounded space.

We assume that the $G$-cells grid of the room containing $N$ sources is expanded into $\mathcal{G}$-cells free space discretization where the actual-virtual sources are active\footnote{If each of the sources have $R$ images, $N(R+1)$ actual-virtual sources are active.}. Given the geometry of the room, the \emph{Image Model} maps the position index $i\in \{1,\ldots, G\}$ of each source to a group $\Omega_{i}\subset \{1,\ldots,\mathcal{G}\}$ containing the location indices of this source and its images (the corresponding virtual sources) in $\mathcal{G}$-points. Consequently, a \emph{free-space} propagation model can be considered between $\mathcal{G}$ actual-virtual source locations and the positions of $M$ microphones. Hence, the forward model between sources and the microphone recordings could be concisely stated  as follows: 
\begin{equation}\label{ReverberationModel}
\mathcal{X} = O P \mathcal{S}. 
\end{equation}
This model holds for each particular independent frequency $f$ of the speech spectrum so we discard the frequency dependency in our mathematical formulation for the sake of brevity.  
Given $\mathcal{X} \in \Cbb^{M\times \mathcal{T}}$, the \emph{observation} matrix of $\mathcal{T}$ frames consisted of spectro-temporal representation of $M$ microphones at a particular frequency band, we decompose the microphone recordings into the following terms:

\begin{itemize}

\item
$\mathcal{S} \in \Cbb^{G\times \mathcal{T}}$ is the \emph{source} matrix whose rows contain $\mathcal{T}$ frames of the spectro-temporal representation of the \emph{actual} sources located in $G$ positions inside the room. Given a fine discretization of the room such that each source occupy an exclusive cell, only $N\ll G$ cells are occupied with active sources and contain nonzero elements and the \emph{support} set $\Scal\subset \{1,\ldots,G\}$ representing the position of those $N$ active sources is sparse. In other words, the \emph{spatial sparsity} indicates $S$ to be a row-sparse matrix with a support corresponding to the position of the actual sources.

\item
$P\in \Rbb_{+}^{\mathcal{G}\times G}$ is the \emph{permutation} matrix such that its $i^{th}$ column contains the absorption factors of $\mathcal{G}$ points on the grid of actual-virtual sources with respect to the reflection of the $i^{th}$ actual source. Since the \emph{Image Model} characterizes the source groups, each column $P_{.,i}$ is consequently supported only on the corresponding group $\Omega_{i}$ i.e., $\forall i\in \{1\ldots, G\}$, $\forall j \notin \Omega_{i}, P_{j,i} = 0$.

\item
$O \in \Cbb^{M\times \mathcal{G}}$ is the \emph{free-space Green's function} matrix such that each $O_{j,i}$ component indicates the sound propagation coefficients, i.e. the attenuation factors and the phase shift due to the direct path propagation of the sound source located at cell $i$ (on a $\mathcal{G}$-point grid of actual-virtual sources) and recorded at the $j^{th}$ microphone. Given the $\mathcal{G}$-cell discretization, $O$ is computed from the propagation formula stated in Equation~(\ref{multipath1}) and it is equal to $\Phi$ when $R=0$.

\end{itemize}

\subsubsection{Source Localization and Absorption Coefficient Estimation}
\label{sec:loc-abs}
Relying on spatio-spectral sparsity of multiple competing sources, the covariance matrix 
of the reverberant recordings exhibits structured sparsity determined by the \emph{Image Model}. We exploit this structured sparsity to identify the location of the active sources and their corresponding absorption coefficients consisting the columns of $P$.
Given the model of the microphone recordings stated in ({\ref{ReverberationModel}), the covariance matrix of the observations is

\begin{eqnarray}
C = \mathcal{X}\mathcal{X}^{*} &=& O \Sigma O^{*} \nonumber\\
&=& \sum_{i=1}^{G} O_{.,\Omega_{i}} \Sigma_{\Omega_{i},\Omega_{i}} O_{.,\Omega_{i}}^{*}\label{cov1},
\end{eqnarray}
where $.^*$ denotes conjugate transpose and $\Sigma =P \mathcal{S} \mathcal{S}^{*}P^{*}$. Note that the spatio-spectral sparsity of concurrent speech sources implies that $\mathcal{S}\mathcal{S}^{*}$ is a diagonal matrix whose diagonal elements specifies the energy of the individual sources - Section \ref{sec:orth} provides some empirical insights on the properties of the covariance matrix. The second equation follows because of the structure of the permutation-attenuation matrix $P$ which indicates that $\Sigma$ is supported only on the set $\bigcup_{i} \Omega_{i}\times \Omega_{i}$ i.e.,
\begin{equation}\label{Sigma}
\begin{split}
 \Sigma_{j,i}=&0 \qquad \forall (j,i)\notin  \bigcup_{i=1}^{G} \Omega_{i}\times \Omega_{i},\\
& \Sigma_{\Omega_{i}, \Omega_{i}} = \|\mathcal{S}_{i,.}\|^{2}_{2}P_{\Omega_{i}.,}P_{\Omega_{i},.}^{*},
\end{split}
\end{equation}
where $\|\mathcal{S}_{i,.}\|_2=\sqrt[2]{\mathcal{S}_{i,.}\mathcal{S}_{i,.}^*}$.  
As we can see, recovering the diagonal elements of $\Sigma_{\Omega_{i}, \Omega_{i}}$ is sufficient to identify the energy of the corresponding source $i$ and the absorption coefficients $P_{\Omega_{i},.}$. We thus focus on recovering these sub-matrices for all $i\in\{1,\ldots, G\}$ from the observation covariance matrix $C$. Using the property of the Kronecker product, we can rewrite \eqref{cov1} as
\begin{equation}\label{Kronecker}
\begin{split}
 & C_{vec} = \underbrace{\begin{bmatrix} B(1) & B(2) & \ldots & B(G)\end{bmatrix}}_{\mathcal{B}}
\underbrace{\begin{bmatrix} v(1) \\ v(2) \\ \vdots \\ v(G)\end{bmatrix}}_{\mathcal{V}}\\ 
\end{split}
\end{equation}

\[
\begin{split}
& \forall i\in \{1\ldots, G\}:\\
& \qquad \qquad \qquad \quad v(i) \defeq \left(\Sigma_{\Omega_{i}, \Omega_{i}}\right)_{vec},\\
& \qquad \qquad \qquad \quad B(i) \defeq \overline O_{.,\Omega_{i}} \otimes O_{.,\Omega_{i}}.
\end{split}
\]
where $\otimes$ denotes the Kronecker product between two matrices and $\overline O_{.,\Omega_{i}}$ is the \emph{element-wise} conjugate of $O_{.,\Omega_{i}}$. In a typical problem setup, very few microphones are used for recording, i.e. $M \ll \mathcal{G} < \sum_{l=1}^{G} |\Omega_{i}| $; thus recovering $\Sigma_{\Omega_{i}, \Omega_{i}}$ requires solving an underdetermined system of linear equations and therefore, in general \eqref{cov1} admits infinitely many solutions and recovery is not feasible.

To circumvent the ill-posedness of the inverse problem, we exploit yet another kind of \emph{block-sparsity} structure that is exhibited in our formulation of the reverberant multi-party recordings. The block sparsity of the actual-virtual sources implies that only $N\ll G$ groups of $v(i)$s (or correspondingly $\Sigma_{\Omega_{i}, \Omega_{i}}$) contain nonzero elements, and thus, identifying those groups equivalently determines the positions of the active sources $\Scal$. In addition, by recovering the corresponding elements of $\mathcal{V}$ and then normalizing them by the sources energies, we can identify the absorption coefficients (i.e., the columns of $P$) which correspond to the attenuation for each source due to the multipath reflections.
   
We simplify the notation by using $\Sigma^i \defeq \Sigma_{\Omega_{i}, \Omega_{i}} \in \Rbb^{|\Omega_i| \times |\Omega_i|}$. Our block-sparse recovery approach can then be formulated by the following  convex minimization problem:

\begin{align}\label{recov}
\argmin_{\Sigma^1,...,\Sigma^{G} } & \quad \sum_{i=1}^{G} \Big\| \Sigma^i_{vec} \Big\|_{L_2} \\
\text{subject to}  & \quad \|C_{vec} - \mathcal{B} \mathcal{V} \|_{L_2} \leq \varepsilon   \nonumber \\
& \quad (\,\, \mathcal{V} = \Big[ (\Sigma^i_{vec})^T, , \ldots, (\Sigma^i_{vec})^T \Big]^T)\nonumber \\
& \quad \Sigma^i= (\Sigma^i)^* \quad \forall i \in \{1,\ldots,G\}  \nonumber \\
& \quad \Sigma^i_{l,j} \geq 0 \quad \qquad \forall l,j,i  \nonumber
\end{align}

We recall that minimizing the sum of the $L_2$ norms of a group of vectors induces the block-sparsity structure in the solution so that, only few subsets of vectors in the group (i.e. few $\Sigma^i$s) contain nonzero elements. Indeed, if $\Sigma^i$s have the same size (i.e. $|\Omega_1| = |\Omega_2| = \ldots =|\Omega_G| $) the objective function of~\eqref{recov} becomes equivalent to the $L_1L_2$ norm\footnote{The $\|.\|_{L_1L_2}$ mixed-norm of a matrix is defined as the sum of the $L_2$ norms of its rows as defined in \eqref{spg_group}} of a matrix whose rows are populated by~$(\Sigma^i_{vec})^T$, which as mentioned earlier is a popular convex approach for block (group) sparse approximation. We solve \eqref{recov} by using the iterative proximal splitting algorithm \cite{Combettesand11}. 

To summarize, we obtain the location of the sources and their images which also corresponds to the support of the room impulse response function for multiple sources. The components of $\Sigma_{\Omega_{i}, \Omega_{i}}$ normalized by the energy of the sources corresponds to the attenuation factors. We entangle the room geometry with the absorption coefficients to characterize the acoustic projections \emph{for any order of desired $R$}, as stated through Equations (\ref{multipath1})-(\ref{dici}). In a scenario where $N < M$, we apply inverse filtering to perform joint speech separation and deconvolution as explained in the following Section \ref{sec:deconv}. \newline 

\subsubsection{Speech Recovery by Inverse Filtering the Acoustic Channel}
\label{sec:deconv}
The approach presented in Sections~\ref{sec:factor} and \ref{sec:loc-abs} enables us to localize the sources and model the mixing channels. Thereby, we can use the frequency domain deconvolution to reverse the attenuation and phase shift induced by the acoustic propagation. Given the frequency domain impulse response function characterized by matrix $H$, we recover the desired signal by inverse filtering stated as

\begin{equation}
 \hat{\mathcal{S}} = (H^TH)^\dag H^T \mathcal{X} 
\label{inverse}
\end{equation}

This operation performs exact deconvolution of the signal from the early room impulse response function~\cite{Huang-Benesty-05,MINT}. The late reverberation can be statistically modeled as an exponentially decaying white Gaussian noise~\cite{book_habet} which also possess the diffuse characteristics~\cite{Cook55}. 

To reduce the effect of late reverberation and enhance the signal, we apply the post-processing proposed in~\cite{PF}. Among several post-filtering methods proposed in the literature~\cite{PF_all, McCowan03}, the Zelinski post-filtering~\cite{PF} is a practical implementation of the optimal Wiener filter; while a precise realization of the later requires knowledge about the spectrum of the desired signal, the Zelinski post-filtering method uses the auto- and cross-power spectra of the multi-channel input signals to estimate the target signal and noise power spectra under the assumption of zero cross-correlation between noise on different sensors. We implemented the Zelinski post-filter for the experiments described in Section~\ref{sec:separation}. The dereverberation of the early impulse response achieved by inverse filtering the acoustic channels enables a more efficient post-filtering as formulated in~\cite{PF}. 

\section{Compressive Acoustic Measurements}
\label{sec:CS_BF}
The approach that we have taken in this paper to address the multi-party speech recovery as studied throughout Sections \ref{sec:msr}-\ref{sec:acoustic}, relies on casting the problem as reconstructing the high-dimensional spatio-spectral information embedded in the acoustic scene from a compressive acquisition provided by the array of microphones. We leveraged model-based sparse recovery theory for characterization of the acoustic measurements and recovering the speech components. In this framework, the theoretical analysis of the performance bounds of our approach is entangled with the performance of the sparse recovery algorithms. A fundamental property to guarantee the theoretical performance bounds is the coherence of the measurement matrix~\cite{TroppWright} defined as

\begin{equation}
 \mu(\Phi) = \max_{1 \leq j,k \leq G, j\neq k} \frac{|\langle \phi_j,\phi_k \rangle|}{\|\phi_j\|\|\phi_k\|}.
\label{mu}
\end{equation}

The coherence quantifies the smallest angle between any pairs of the columns of $\Phi$. The number of recoverable non-zero coefficients ($K$) using either convexified or greedy sparse recovery is inversely proportional to $\mu$ \cite{TroppWright} as
\[
 K < \frac{1}{2}(\mu^{-1}+1) 
\]

Hence, to guarantee the performance of sparse recovery algorithms, it is desired that the coherence is minimized. As the measurement matrix is constructed of the location-dependent projections, this property implies that the contribution of the source to the array's response is small outside the corresponding sensor location or equivalently the resolution of the array is maximized. It has been shown in~\cite{Carin1} that the free-space Green's function constituted projections given that the inter-element spacing is large enough exhibit an optimal design and the columns of the measurement matrix corresponds to a sampled Fourier basis function. It has been further pointed out that a large-aperture random design of sensor array yields the projections to be mutually incoherent~\cite{Carin1,Carin2}. Thereby the projections are spread across all the acoustic scene and each sensor captures the information about all components of $\mathcal{S}$. These studies elucidate that the performance of our sparse approximation framework is entangled with the microphone array construction design. This issue is addressed in Section~\ref{sec:tests}.

\section{Experimental Analysis}
\label{sec:tests}

\subsection{Data Recordings Set-up}
\label{sec:monc}
Experiments were performed in the framework of the Multichannel Overlapping Numbers Corpus (MONC). This database is acquired by playback of utterances from the Numbers Corpus release 1.0, prepared by the Center for Spoken Language Understanding at the Oregon Graduate Institute~\cite{MONC}. 

The recordings were made in a 8.2m $\times$ 3.6m $\times$ 2.4m rectangular room containing a centrally located 4.8m $\times$ 1.2m rectangular table. The positioning of loudspeakers was designed to simulate the presence of 3 competing speakers seated around a circular meeting room table of diameter 1.2m. The loudspeakers were placed at 90±$^{\circ}$ spacings at an elevation of 35cm (distance from table surface to center of main speaker element). An eight-element, 20cm diameter, circular microphone array placed in the center of the table recorded the mixtures. The recording scenario is illustrated in Fig. \ref{fig:monc}. One hour of speech signals are recorded at 8 kHz sampling frequency. The average signal to noise ration (SNR) of the recordings is 9dB.

\begin{figure}[t]
  \centering
  \includegraphics[width=0.55\columnwidth]{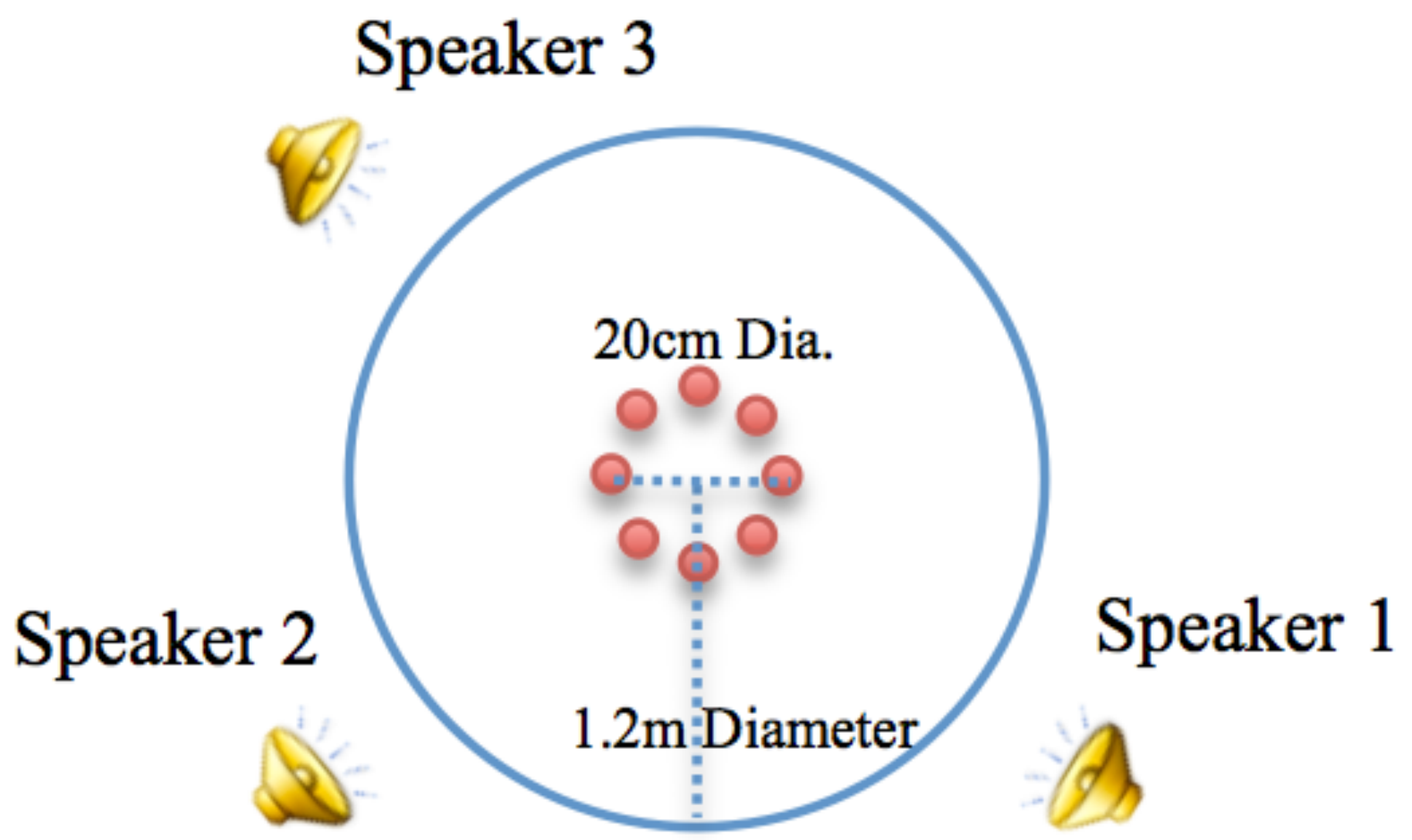}
  \caption{\small{Loudspeaker and microphone placement used for recording MONC corpus \cite{MONC}.}}
  \label{fig:monc}
\end{figure}

\subsection{Orthogonality of Spectrographic Speech}
\label{sec:orth}
We carried out experiments to investigate the orthogonality of multiple speech sources in Fourier domain. In this experiment, 3 speech signals, 9s each, is analyzed in frames of size 128ms (fft-size = 1024) with 50\% overlap; thus we obtain 3 matrices of 512 by 140 corresponding to the STFT of each source. The orthogonality is measured for each frequency band independently. We construct the matrix $\mathcal{X}_{3 \times 140}$ where each row corresponds to each source and has the frequency components of a particular band along 140 frames. In case of perfectly orthogonal sources, $C = \mathcal{X}\mathcal{X}^*$ is Identity and the energy of the diagonal of the matrix is equal to the matrix Frobenius norm. Fig.~\ref{fig:speech_orth}-right-hand-side illustrates the diagonal-$L_2$-norm$/$matrix-Frobenius-norm. 

In addition, we performed some experiments by pointwise multiplication of the STFTs of two utterances and plot the histograms of the resulted values. Fig.~\ref{fig:speech_orth}-left-hand-side illustrates the obtained histogram. As we can see the distribution mass of the energy of the point-wise multiplication values is localized around $0$. This phenomenon indicates that the majority of the high energy components in spectro-temporal domain are non-overlapping or disjoint.  

\subsection{Room Geometry Estimation}
\label{sec:geom-exp}
The first step to characterize the room acoustic is to estimate the room geometry. We accomplish this step through localization of the images of multiple speakers in a large extended area using the sparse recovery framework with a free space model as followed by the least-squares regression of the room geometry as explained in Section \ref{sec:geom}. 

The location of the source images corresponds to the support of the room impulse response function. The energies of the recovered signals are sorted and truncated to the order of $D(D+1)/2$ where $D$ denotes the number of reflective surfaces and it is equal to 6 in our study to cover the support of the early reflections of the walls to guarantee the uniqueness of the solution~\cite{hearRoom}. The estimated support of the room impulse response function is then used for estimation of the room rectangular geometry by generating the room impulse responses for various room geometries and identify the best fit to the estimated support in least-squares sense.  

\if@twocolumn
  \begin{figure}[!ht]
    \includegraphics[height=30.7mm,width=0.24\textwidth]{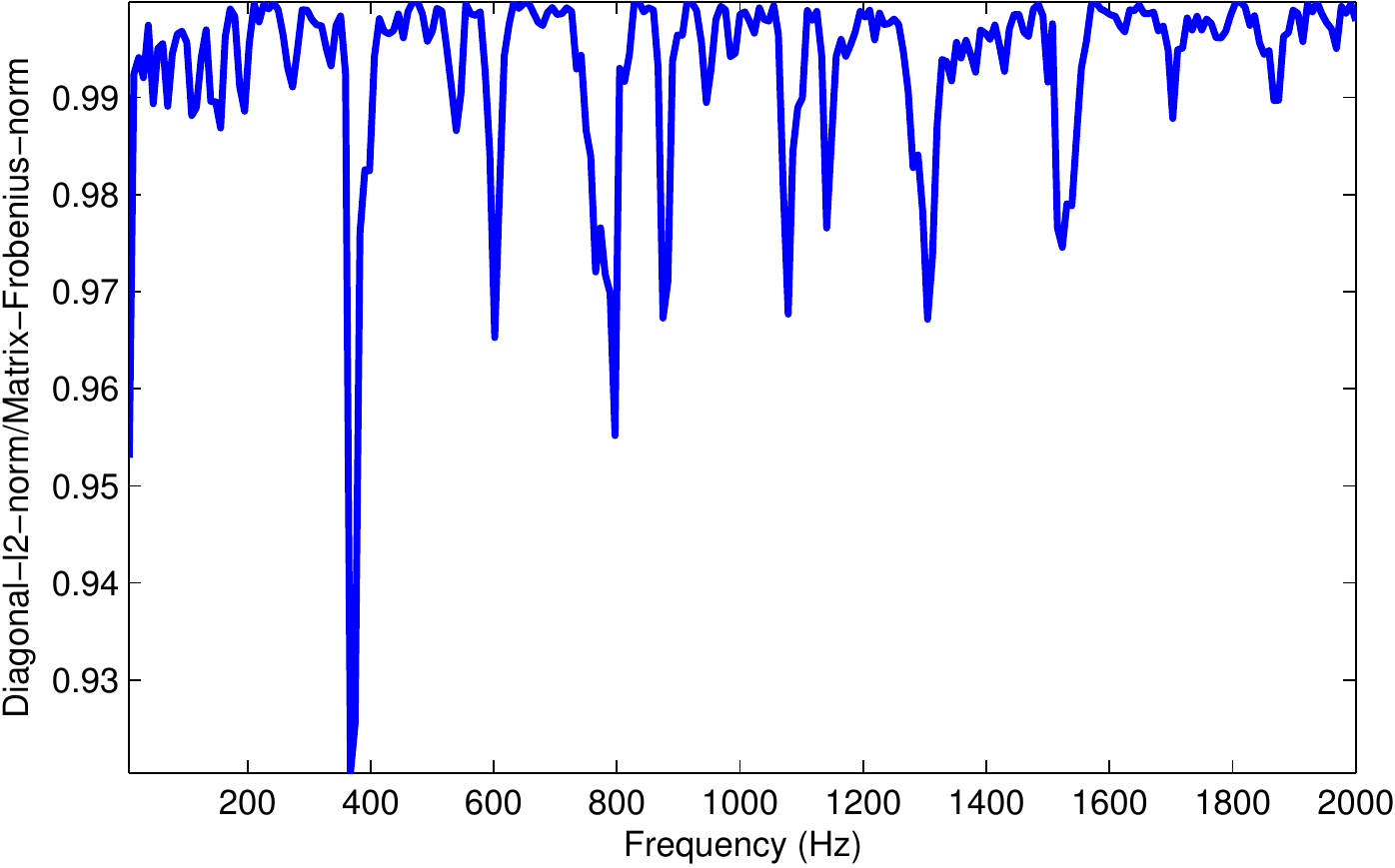}
    \includegraphics[height=31mm,width=0.24\textwidth]{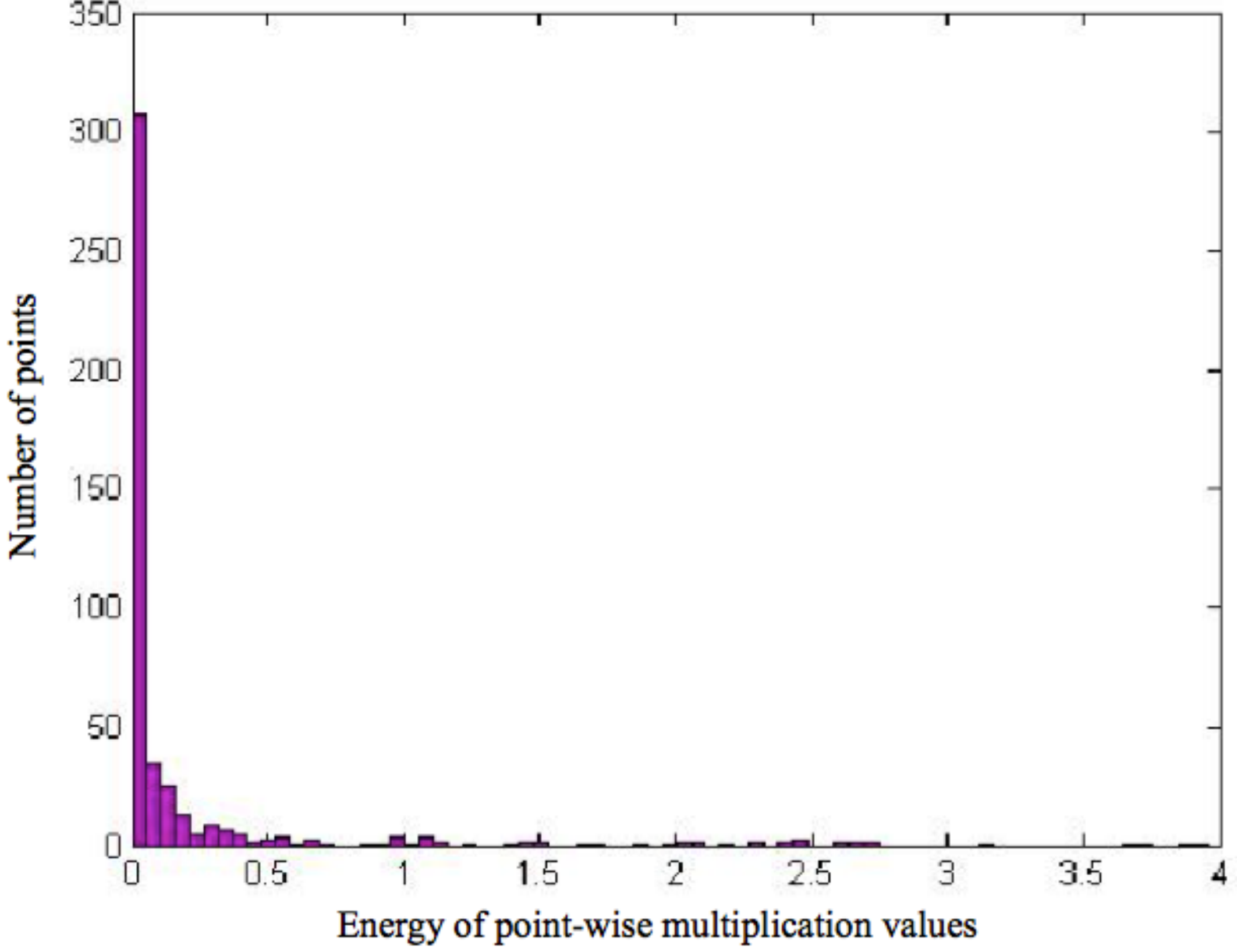}
    \caption{\small{Orthogonality of multiple speech utterances in spectro-temporal domain: Right-hand-side illustrates the diagonal-$L_2$-norm$/$matrix-Frobenius-norm of the covariance matrix constructed per frequency and Left-hand-side illustrates the energy histogram of the component-wise multiplication of speech utterances.}}
    \label{fig:speech_orth}
  \end{figure}
\else
  \begin{figure}[t!]
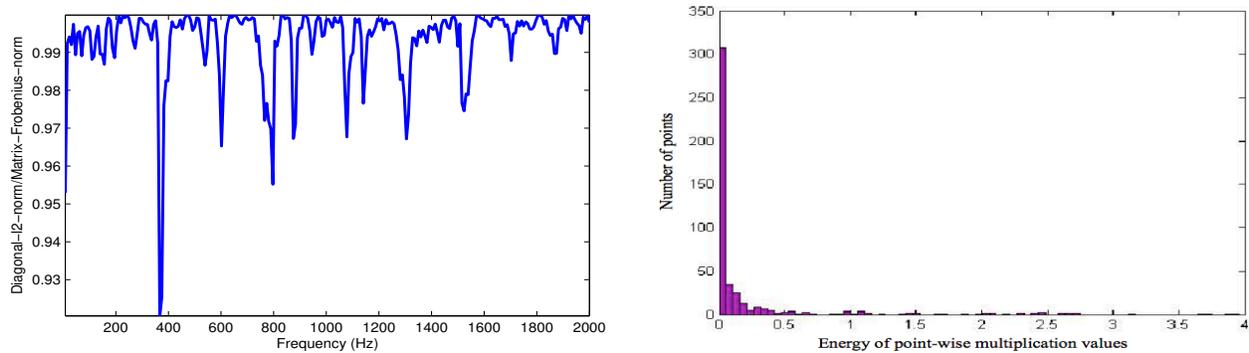

    \includegraphics[height=4.5cm, width=0.48\textwidth]{Fig/speech_orthogonality_small-crop}  
    \hfill
    \includegraphics[height=4.6cm, width=0.48\textwidth]{Fig/pdf-hist}  
    \caption{Speech orthogonality}
    \label{fig:speech_orth}
    \end{figure}
\fi

The planar area of the room is divided into cells with 25cm spacing. The distance threshold to identify the actual sources is selected as 1m. To achieve a higher estimation, we restricted our discretized gird to the orthogonal subspaces corresponding to the orthogonal walls. We could estimate the geometry of the room up to 50cm error from the recordings of 3 sources in a close proximity to the microphone array as depicted in Fig. \ref{fig:monc}.

\subsection{RIR Estimation}
The second step to characterize the room acoustic is to estimate the absorption coefficients of the reflective surfaces. We accomplish this step through estimation of the room impulse response (RIR) function by implementing the technique explained in Section~\ref{sec:RIR}. We used the CVX software package~\cite{CVX} for optimization formulated in (\ref{cr5}) while sigma is chosen 0.1. The data was provided by concatenating 20 single speaker speech utterances.

The super-resolution source localization is performed based on the energy recovered from each cell using sparse recovery framework while the forward model corresponds to the direct path propagation and the support of the RIR function was determined considering a 6-sided model of an enclosure with the known geometry. We assumed that the reflections of the carpet floor are trapped under the table; hence, the meeting table was considered as the floor in our Image Model. The room reverberation time is measured about 100 ms from the energy decay curve of the estimated RIR and the reflection coefficients are estimated as 0.1 for the walls as well as the ceiling and 0.6 for the meeting table. Our estimation matches the empirical Sabin-Franklin's formula~\cite{Habets}:  

\begin{equation}
RT_{60} = \frac{24 \ln(10) V}{c \sum_{i=1}^6 W_i(1-\iota_i^{2})},
\label{beta}
\end{equation}
where $V$ denotes the volume of the room, $\iota_i$ the reflection coefficient, and $W_i$ the surface of the $i^{th}$ wall. 

Although our method is blind, we verified the estimated impulse response and the corresponding reflection coefficients through adaptive filtering technique using the original clean speech provided at MONC from the original Numbers corpus. 
Figure~\ref{rir} shows the effectiveness of the room impulse response estimation with the structured sparsity constraints and the alternative least-squared optimization from noisy data. 

\begin{figure}[t]
  \centering
  \includegraphics[width=\columnwidth]{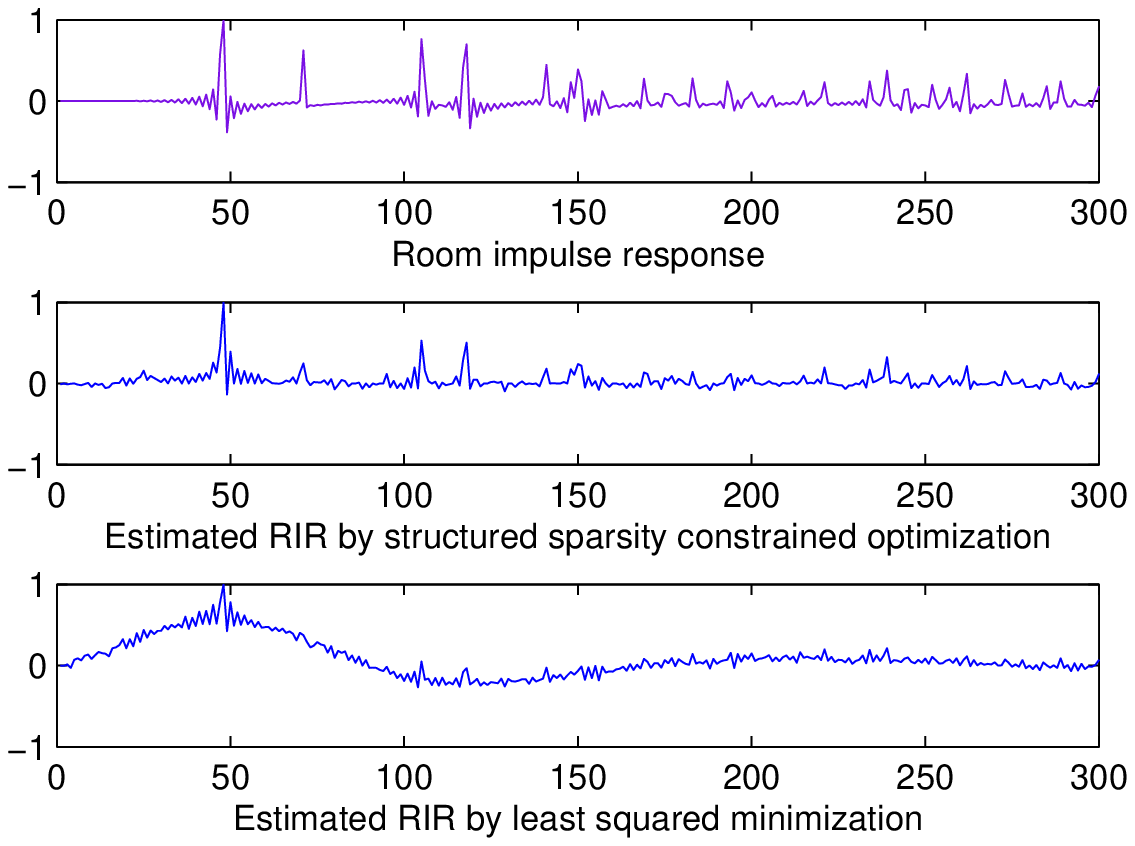}
  \caption{\small{Room Impulse Response (RIR) estimation from noisy measurements}}
  \label{rir}
\end{figure}

\subsection{Structured Sparse Speech Recovery Performance}
\label{sec:separation}
The speech recovery experiments are performed using different sparse recovery approaches to incorporate the block inter-connection as well as harmonicity of the spectro-temporal coefficients of speech signal. The spectro-temporal representation required for speech recovery is obtained by windowing the signal in 256ms frames using a Hann function with 25\% overlap. The quality evaluation results in terms of Signal to Interference Ratio (SIR)~\cite{bss_eval} and Perceptual Evaluation of Speech Quality (PESQ)~\cite{persia} are summarized in Figure~\ref{fig:bar_random_SR}. The block-size $b$ was set to 4 as it was shown yielding the best results, especially for B-OMP and B-$L_1L_2$. 

In the harmonic model, we consider that $f_0 \in [150-400]$ Hz. Those frequencies that are not the harmonics of $f_0$ are recovered independently in H-IHT and H-$L_1L_2$. We also considered that the harmonic structures are non-overlapping and $k$ spans the full frequency band. For H-OMP, the harmonic subspaces are used to select the bases while projection is performed for the full frequency band.

We observe that the highest quality in terms of SIR and PESQ are obtained by convex optimization. This could be due to the zero-forcing spirit of greedy approaches. This deficiency is particularly exhibited for speech-like signals, which do not possess high compressibility~\cite{asaei3}. However, in some applications such as speech recognition, where the reconstruction of the signal is not required, we can exploit the sparsity of the information bearing components in greedy sparse recovery approaches, which offer a noticeable computational speed in efficient implementations~\cite{V_CAMSAP, ThomasMike} and a reasonable performance~\cite{asaei2}. 

Considering the speech signal model consisted of voiced and unvoiced segments, the block-interdependency mostly corresponds to the unvoiced speech while the harmonicity is exhibited in the voiced segments; hence we expect that a combination of both of the structures is beneficial for efficient speech recovery. 

Comparing the results with the conventional uniform-array, we observe that the random setting of microphone array can significantly improve the quality of the separated speech. Hence, the compact uniform microphone array set-up is not an optimal design from the sparse reconstruction standpoint and the present study motivates more investigation on sparse and ad-hoc microphone array layouts. 

\begin{figure}[t]
  \centering
  \includegraphics[width=\columnwidth]{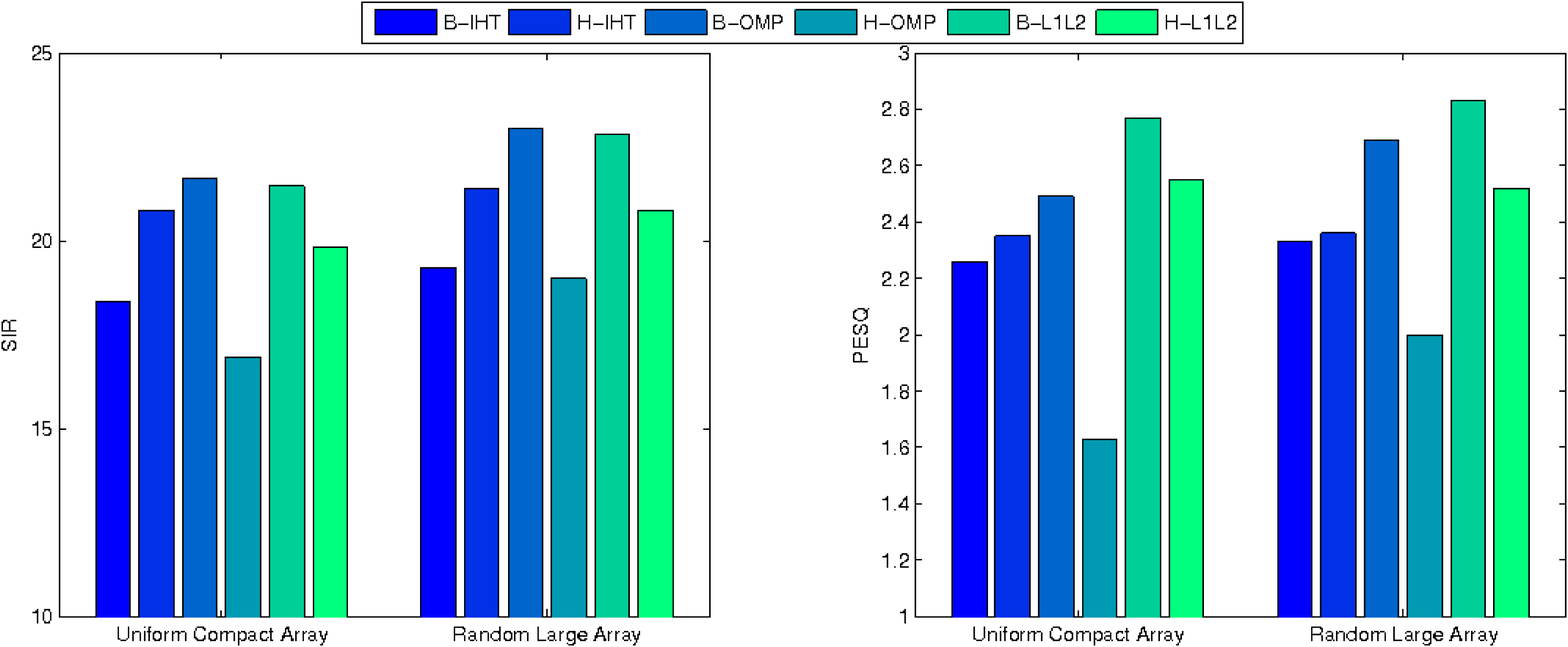}
  \footnotesize
  \caption{\small{Quality evaluation of the separated speech using different sparse recovery approaches in terms of SIR and PESQ. The baseline measures are -3.68 and 1.44 respectively}}
  \label{fig:bar_random_SR}
\end{figure}

\subsection{Multi-party Acoustic Modeling and Speech Recovery}
\label{sec:separation}
\subsubsection{Synthetic data evaluations}
We perform some initial evaluations on synthetic data in various noisy and reverberant conditions to validate our approach explained in Section \ref{sec:absorption}. The results of these experiments elucidate the empirical performance bounds for absorption coefficient estimation and signal recovery using block sparse recovery algorithm. 

We consider the following recording set-ups: (1) 8-channel circular microphone array positioned in the middle of the room, (2) 12-channel microphone array: two sets of 6-channel circular array located far apart, (3) 16-channel microphone array: two sets of 8-channel circular array located far apart. We considered about 3cm displacement of the microphones. Evaluations are carried out using 1-3 sources distributed arbitrarily in the room with the following characteristics (a) Spectrum of orthogonal random broad-band sources at 52 auditory-centered frequencies and (b) Spectrum of independent speech sources at the frequency-bands which contain 80\% of the total energy. The results of source localization (SL), absorption coefficients estimation (AC) and signal recovery (SR) are illustrated in Figs. (\ref{fig:bar_green}) and (\ref{fig:bar_speech}). 

\if@twocolumn
  \begin{figure*}[!ht]
    \begin{minipage}{0.49\linewidth}
      \includegraphics[width=\figwidth]{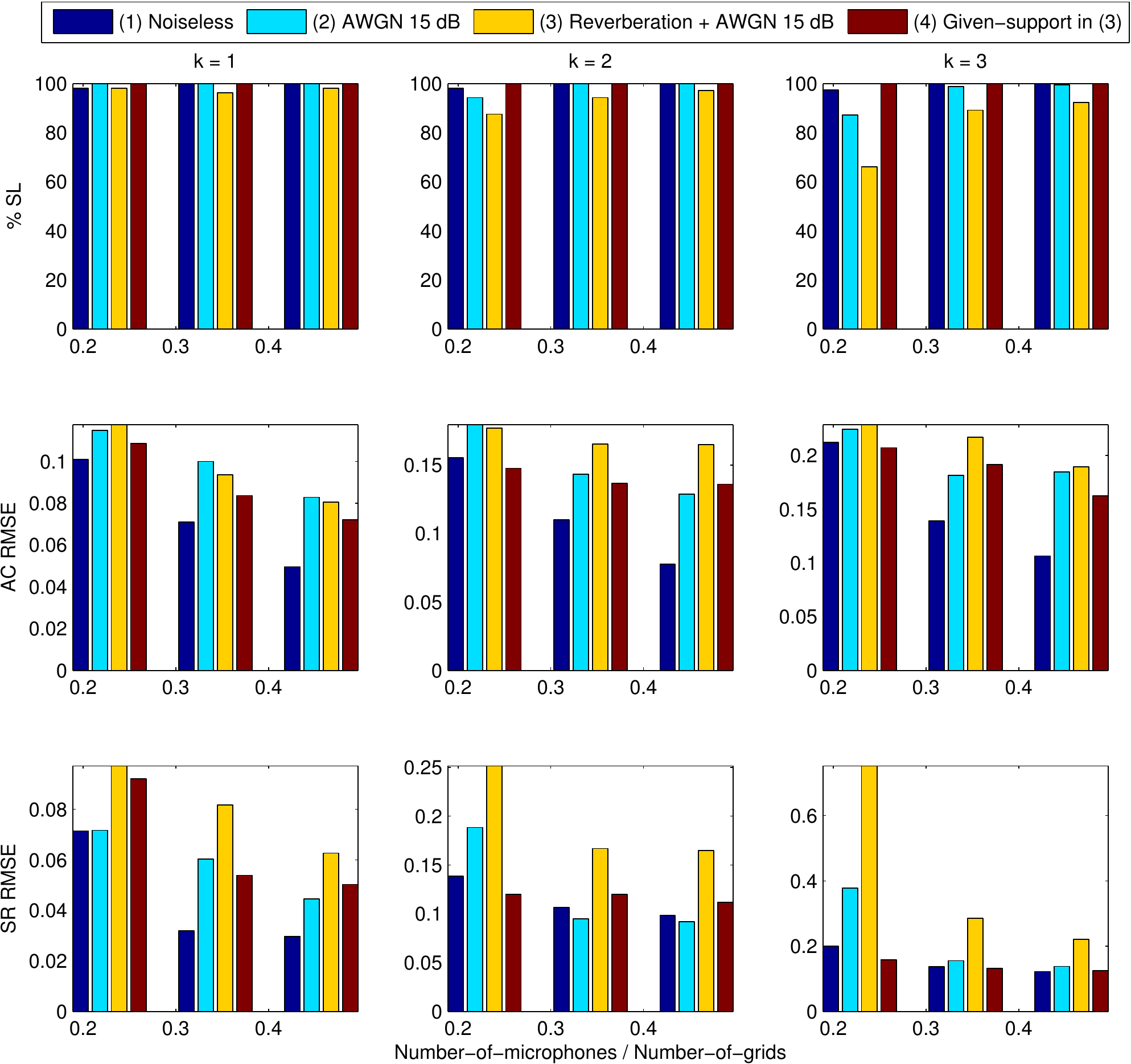}
      \caption{\small{Performance of the algorithm in terms of Source Localization (SL), Root Mean Squared Error (RMSE) of Absorption Coefficients (AC) estimation as well as Signal Recovery (SR). The test data are random orthogonal sources and the measurement matrix consisted of the free-space Green's function.}}\label{fig:left}
      \label{fig:bar_green}      
    \end{minipage}
    \hfill
    \begin{minipage}{0.49\linewidth}
      \includegraphics[width=\figwidth]{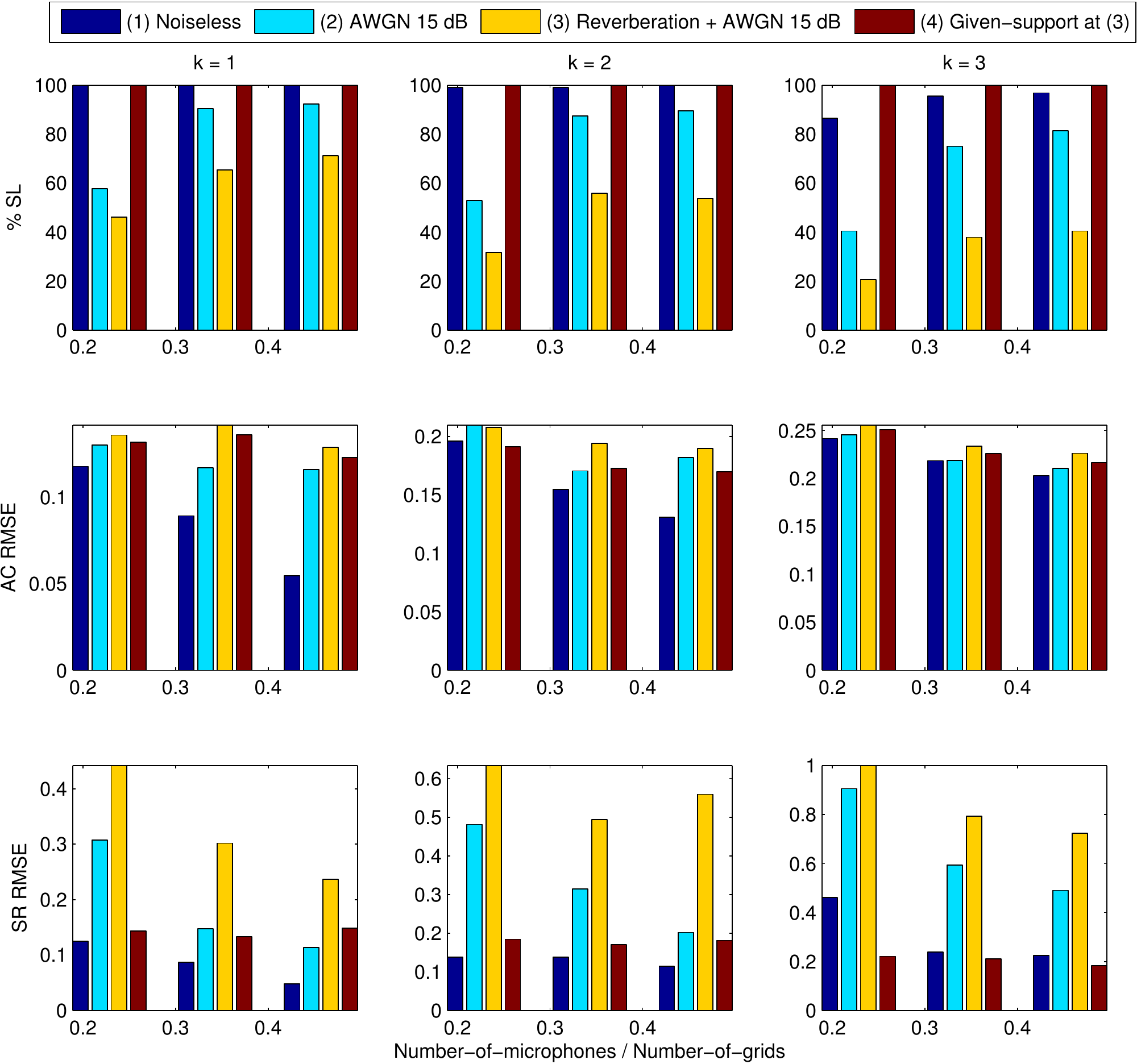}
      \caption{\small{Performance of the algorithm in terms of Source Localization (SL), Root Mean Squared Error (RMSE) of Absorption Coefficients (AC) estimation as  well as Signal Recovery (SR). The test data are random speech utterances and the measurement matrix consisted of the free-space Green's function.}}\label{fig:right}
      \label{fig:bar_speech}
    \end{minipage}
     \end{figure*}
\else
  \begin{figure*}[t!]
    \begin{minipage}{0.49\linewidth}
      \includegraphics[width=\figwidth]{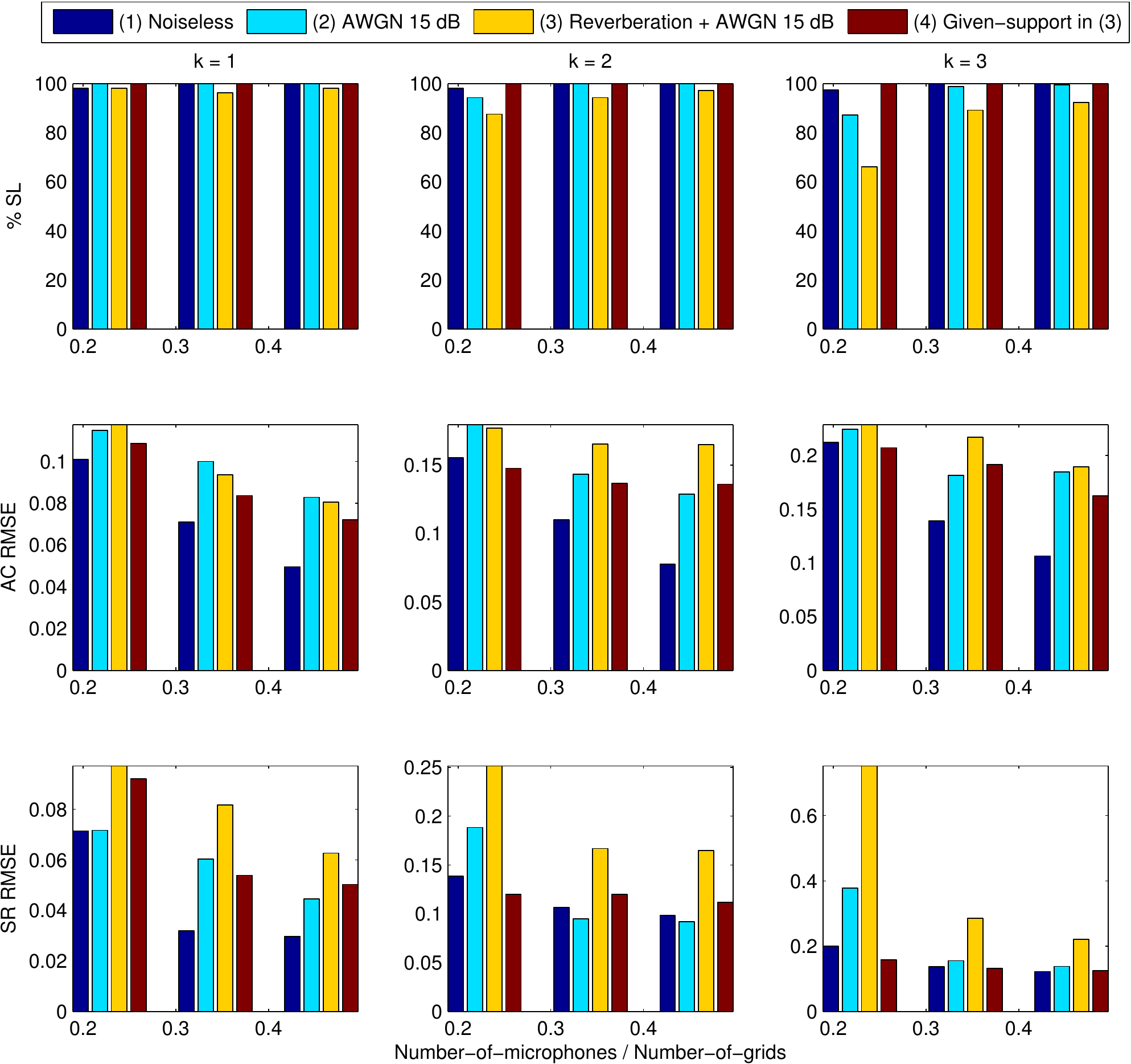}
      \caption{\small{Performance of the algorithm in terms of Source Localization (SL), Root Mean Squared Error (RMSE) of Absorption Coefficients (AC) estimation as well as Signal Recovery (SR). The test data are random orthogonal sources and the measurement matrix is the free-space Green function}}\label{fig:left}
      \label{fig:bar_green}      
    \end{minipage}
    \hfill
    \begin{minipage}{0.49\linewidth}
      \includegraphics[width=\figwidth]{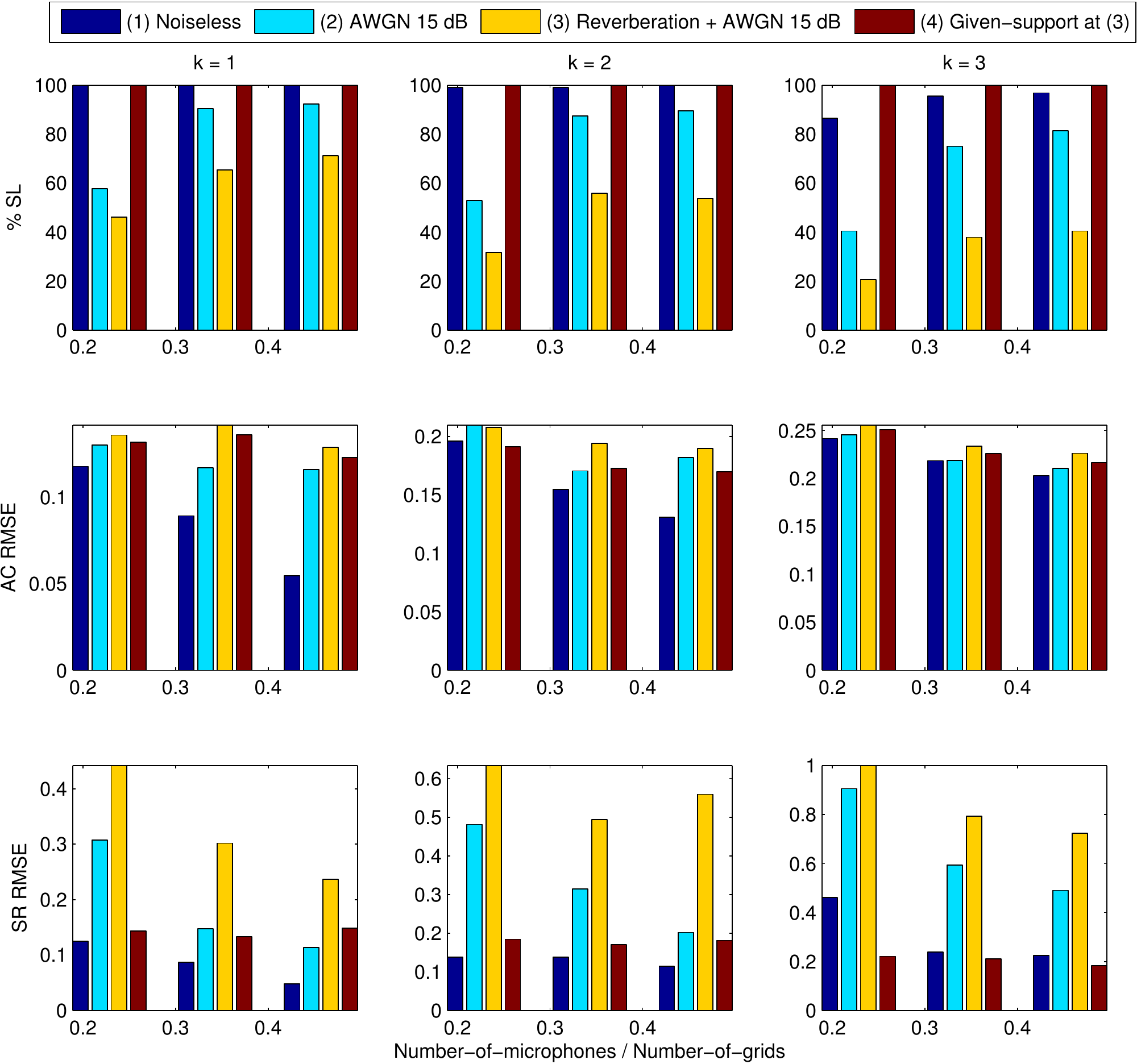}
      \caption{\small{Performance of the algorithm in terms of Source Localization (SL), Root Mean Squared Error (RMSE) of Absorption Coefficients (AC) estimation as  well as Signal Recovery (SR). The test data are random speech utterances and the measurement matrix is the free-space Green function}}\label{fig:right}
      \label{fig:bar_speech}
    \end{minipage}
     \end{figure*}
\fi

\subsubsection{Speech Recovery Performance}
Given the location of the sources and the characterized room acoustic channel, we recover the desired signal by inverse filtering and perform speech recognition. 

The automatic speech recognition (ASR) scenario was designed to broadly mirror that of Moore and McCowan~\cite{Mccowan_overlapping}. A typical front-end was constructed using the HTK toolkit \cite{HTK} with 25ms frames at a rate of 10ms. This produced 12 mel-cepstra plus the zero$^{th}$ coefficient and the first and second time derivatives; 39 features in total. Cepstral Mean Normalization (CMN) is applied to the feature vectors, resulting in speech recognition performance improvement of about 15\% relative. The ASR accuracy on the clean speech data is about 95\%. We perform MAP adaptation by training directly on recovered data. The Zelinsky post-filtering is applied on the recovered speech prior to the recognition~\cite{PF}.

In addition to the speech recognition, we evaluate the quality of the recovered speech using SIR ~\cite{bss_eval} as well as PESQ~\cite{persia}. As our methods rely on the principles of spatial diversity, we compare them with beamforming techniques which possess similar essence. We used the super-resolution speaker localization based on sparse recovery to perform near-field beamforming. In addition, we compared our method with the sparse RIR-estimation approach described in Section \ref{sec:RIR} which relies on least-squares fitting of the Image Model (RIR-LS)~\cite{asaei3} for estimation of the absorption factors. The resulting speech recovery performance is summarized in Table~\ref{results1}.

As the results indicate, the proposed RAM-SR method yields the maximum interference suppression and highest perceptual quality of the recovered speech in multi-party scenarios as quantified in terms of SIR and PESQ. It also outperforms other techniques in terms of word recognition rate.

The results support importance of the structured sparsity models to recover the spatio-spectral information from the multi-channel recordings. The spectral dependencies of the speech components could be further parametrized through auto-regressive (AR) models where we could characterize the dependencies along the temporal sequences or oblique structures in spectro-temporal domain \cite{Ellis07}. We will further incorporate these structures into the framework of sparse recovery to devise the speech-specific algorithms which enable more efficient recovery performance. We could further exploit the statistical dependencies \cite{Elad12} for the specific task of speech recognition. The approach that presented to model the acoustic channel relies on joint-sparse recovery. Give the low-rank structure of the problem induced by the similar signals attributed to the source and its images, a promising extension of this work would be exploiting the low-rank and joint-sparse recovery algorithms as we studied in \cite{mohammad12}. 

\subsubsection{Real data evaluations}
The scenario of the real data tests is explained in Section \ref{sec:monc} which is similar to the first set-up described above. We assume the location of the desired source to be fixed throught out the whole session. The estimated absorption coefficients are plotted using the data in the following conditions: (I) single speech utterances, (II) Two simultaneous speech utterances, (III) Three simultaneous speech utterances. The estimates are run over 9000 speech files of MONC corpus~\cite{MONC} and computed and averaged for each frequency-band independently. The estimated absorption coefficients for each frequencies (computed at a resolution of 4 Hz) are illustrated in Fig.~\ref{fig:AC}.

\begin{table} [t]
\tiny
\footnotesize
  \caption{\small{Quality evaluation of the recovered speech in terms of Source to Interference Ratio (SIR), Perceptual Evaluation of Speech Quality (PESQ) and Word Recognition Rate (WRR) using near-field Super Directive (SD) beamforming, vs. inverse filtering of the RIR estimation based on the least-squared estimation of the absorption factors (RIR-LS) and the proposed Room Acoustic Modeling via block Sparse Recovery (RAM-SR)}}
\label{results1}
\centerline{
\begin{tabular}{|c|c|c|c|c|c|c|}
\hline
 N & Meas. & Baseline & Lapel & SD & RIR-LS & RAM-SR \\
\hline
\hline
\multirow{3}{*}{1} & SIR & 12.3 & 19.19 & 18.52 & 16.5 & 16.1 \\ 
 & PESQ & 2.7 & 3 & 3.3 & 2.91 & 2.97 \\
 & WRR\% & 89.61 & 93.21 & 95 & 93.67 & 93.3 \\
\hline
\multirow{3}{*}{2} & SIR & 2.6 & 18.29 & 11.33 & 12.5 & 17.5 \\ 
 & PESQ & 2 & 2.35 & 2.69 & 2.6 & 2.8 \\
 & WRR\% & 55.19 & 74.53 & 68.16 & 83.37 & 87.93 \\
\hline
\multirow{3}{*}{3} & SIR & -0.7 & 18.35 & 10 & 10.1 & 14.2 \\ 
 & PESQ & 1.6 & 2.27 & 2.48 & 2.4 & 2.62 \\
 & WRR\% & 39.92 & 68.13 & 61.45 & 70.88 & 79.21 \\
\hline
\end{tabular} }
\end{table}

\begin{figure}[h]
  \centering
  \includegraphics[width=0.9\figwidth]{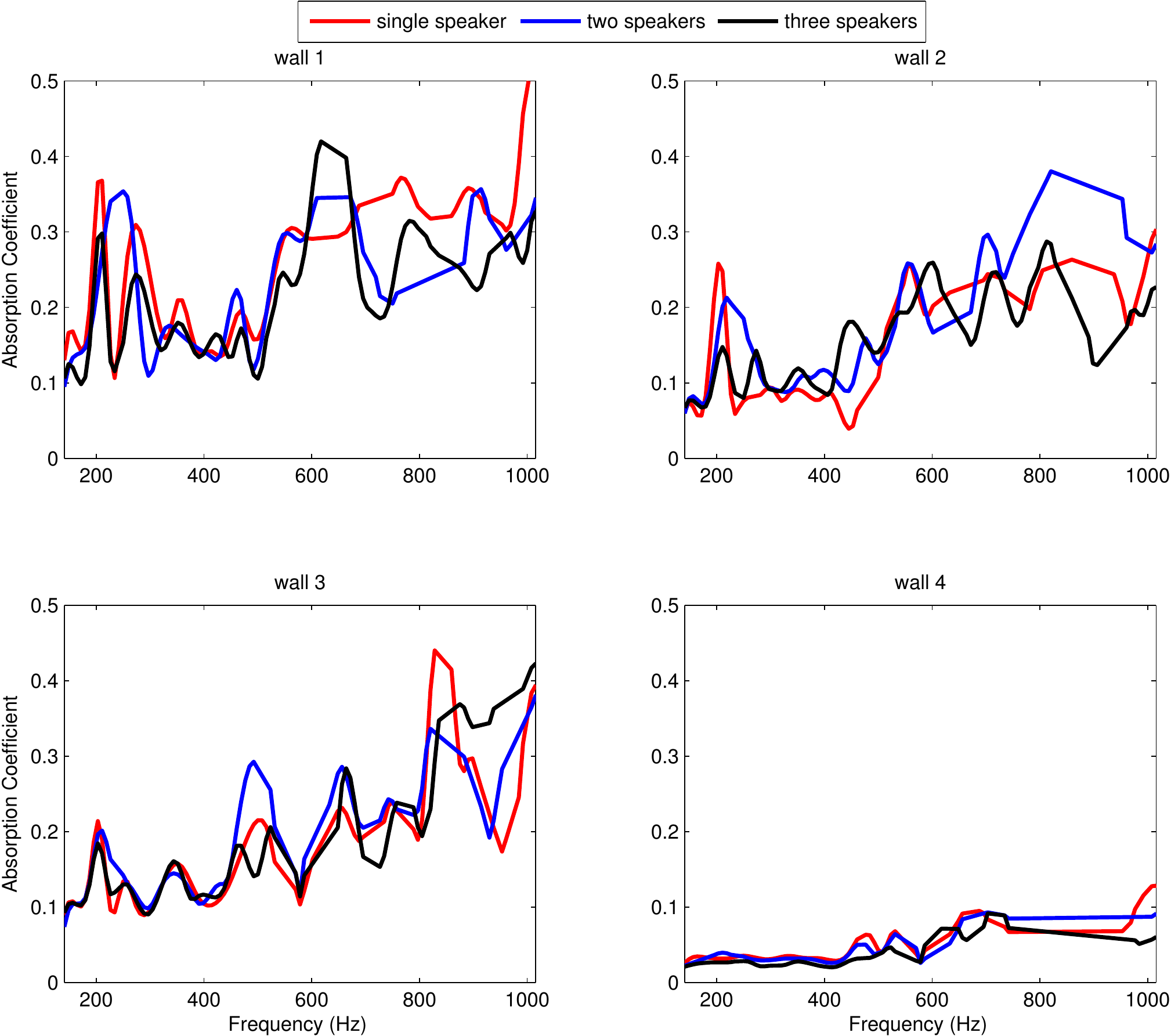}
  \footnotesize
  \caption{{\small{Frequency-dependent absorption coefficients computed for each wall from the utterances of 3 competing speakers for the third speaker.}}}
  \label{fig:AC}
\end{figure}

\section{Conclusions}
\label{sec:final}

We addressed recovery of the speech information from structured sparse reconstruction perspective where we exploited spatial, spectral as well as acoustic structures underlying the representation of multiparty reverberant recordings to characterize the measurements and to recover the individual signals. 

Relying on the structured sparsity of Image model of multipath effect, we identified the acoustic channel through (1) estimation of the room geometry by localization of the sources and low-rank clustering of the subspaces corresponding to each source and (2) estimation of the absorption coefficients using block-sparse recovery algorithm. Given the acoustic channel, we characterized the compressive acoustic projections and cast multiparty speech recovery as structured sparse approximation where we exploited the block dependency as well as harmonicy of the spectral coefficients to recover the speech signals. In addition, we showed that in a well-defined set-up we can pereform joint separation and deconvolution by frequency domain inverse filtering of the acoustic channels. 

The proposed theory is validated by quantitative assessments performed through extensive experiments. The results on real data recordings demonstrate the applicability of our method for recovery of speech in multi-party scenarios for higher level application of distant speech recognition. The results motivate incorporating further parametrized sparsity structures to devise speech-specific recovery algorithms. They also support construction designs relying on ad-hoc and sparse microphone arrays layout for efficient capturing and extraction of the information embedded in the acoustic scene.

\section*{Acknowledgment}
The authors would like to thank Prof. Bhiksha Raj from Machine Learning for Signal Processing (MLSP) group at Carnegie Mellon University for the valuable comments and fruitful discussions which resulted in important improvements.

The research leading to these results has received funding from the European Union under the Marie-Curie Training project SCALE (Speech Communication with Adaptive LEarning), FP7 grant agreement number 213850.

VC acknowledges Rice University for his Faculty Fellowship, MIRG-268398, ERC Future Proof, DARPA KeCoM program \#11-DARPA-1055 and SNF 200021-132548. 

\ifCLASSOPTIONcaptionsoff
  \newpage
\fi

\footnotesize{
\bibliographystyle{IEEEtran}
\bibliography{refs,strings}

\begin{thebibliography}{10}
\providecommand{\url}[1]{#1}
\csname url@samestyle\endcsname
\providecommand{\newblock}{\relax}
\providecommand{\bibinfo}[2]{#2}
\providecommand{\BIBentrySTDinterwordspacing}{\spaceskip=0pt\relax}
\providecommand{\BIBentryALTinterwordstretchfactor}{4}
\providecommand{\BIBentryALTinterwordspacing}{\spaceskip=\fontdimen2\font plus
\BIBentryALTinterwordstretchfactor\fontdimen3\font minus
  \fontdimen4\font\relax}
\providecommand{\BIBforeignlanguage}[2]{{%
\expandafter\ifx\csname l@#1\endcsname\relax
\typeout{** WARNING: IEEEtran.bst: No hyphenation pattern has been}%
\typeout{** loaded for the language `#1'. Using the pattern for}%
\typeout{** the default language instead.}%
\else
\language=\csname l@#1\endcsname
\fi
#2}}
\providecommand{\BIBdecl}{\relax}
\BIBdecl

\bibitem{SiSEC2011}
S.~Araki, F.~Nesta, E.~Vincent, Z.~Koldovsky, and G.~Nolte, ``The 2011 signal
  separation evaluation campaign (sisec2011): Audio source separation,'' vol.
  7191, 2011.

\bibitem{pas1}
E.~Shriberg, A.~S. A., and D.~Baron, ``Observations on overlap: Findings and
  implications for automatic processing of multi-party conversation,'' in
  \emph{In Proceedings of Eurospeech}, 2001.

\bibitem{Ozerov-12}
A.~Ozerov, ``A general flexible framework for the handling of prior information
  in audio source separation,'' vol. 20(5), 2012.

\bibitem{Douglas-05}
S.~C. Douglas, H.~Sawada, and S.~Makino, ``Natural gradient multichannel blind
  deconvolution and speech separation using causal fir filters,'' vol.~13,
  2005.

\bibitem{TRINICON}
H.~Buchner, R.~Aichner, and W.~Kellermann, ``{TRINICON}-based blind system
  identification with application to multiple-source localization and
  separation,'' vol.~13, 2007.

\bibitem{BSS}
S.~Makino, T.~Lee, and H.~Sawada, ``Blind speech separation,'' \emph{Springer},
  2007.

\bibitem{mike_geom}
M.~A. Dmour and M.~E. Davies, ``A new framework for underdetermined speech
  extraction using mixture of beamformers,'' \emph{{IEEE} Transactions on
  Audio, Speech, and Language Processing}, vol.~19, pp. 445--457, 2011.

\bibitem{Araki-07}
S.~Araki, H.~Sawada, and S.~Makino, ``Blind speech separation in a meeting
  situation,'' in \emph{In Proceedings of ICASSP}, 2007.

\bibitem{DSR-book}
M.~Wolfel and J.~McDonough, ``Distant speech recognition,'' 2009.

\bibitem{spars1}
M.~Zibulevsky and B.~A. Pearlmutter, ``Blind source separation by sparse
  decomposition in a signal dictionary,'' vol. 13(4), 2001.

\bibitem{Remi}
R.~Gribonval and S.~Lesage, ``A survey of sparse component analysis for blind
  source separation: Principles, perspectives, and new challenges,'' in
  \emph{ESANN, 14th European Symposium on Artificial Neural Networks}, 2006.

\bibitem{Zibulevsky}
P.~Bofill and M.~Zibulevsky, ``Underdetermined blind source separation using
  sparse representations,'' \emph{Signal Processing}, 2001.

\bibitem{ellq}
R.~Saab, O.~Yilmaz, M.~J. Mckeown, and R.~Abugharbieh, ``Underdetermined
  anechoic blind source separation via \(\ell_q\)-basis-pursuit with \(q<1\),''
  \emph{{IEEE} Transactions on Signal Processing}, 2007.

\bibitem{Nesta-12}
F.~Nesta and M.~Omologo, ``Convolutive underdetermined source separation
  through weighted interleaved ica and spatio-temporal source correlation,''
  vol. 7191, 2012.

\bibitem{Huang-Benesty-05}
Y.~Huang, J.~Benesty, and J.~Chen, ``A blind channel identification-based
  two-stage approach to separation and dereverberation of speech signals in a
  reverberant environment,'' vol. 13(5), 2005.

\bibitem{Huang-Benesty-03}
Y.~Huang and J.~Benesty, ``A class of frequency-domain adaptive approaches to
  blind multichannel identification,'' vol. 51(1), 2003.

\bibitem{MINT}
M.~Miyoshi and Y.~Kaneda, ``Inverse filtering of room acoustics,'' \emph{{IEEE}
  Transactions on Audio, Speech, and Language Processing}, 36(2), 1988.

\bibitem{Rotili-10}
R.~Rotili, C.~D. Simone, A.~Perelli, A.~Cifani, and S.~Squartini, ``Joint
  multichannel blind speech separation and dereverberation: A real-time
  algorithmic implementation,'' in \emph{In Proceedings of 6th International
  Conference on Intelligent Computing}, 2010.

\bibitem{Yoshioka-10}
T.~Yoshioka, T.~Nakatani, M.~Miyoshi, and H.~Okuno, ``Blind separation and
  dereverberation of speech mixtures by joint optimization,'' vol. 19(1), 2010.

\bibitem{Nakatani-11}
T.~Nakatani, T.~Yoshioka, and K.~Kinoshita, ``Mathematical analysis of speech
  dereverberation based on time-varying gaussian source model: Its solution and
  convergence characteristics,'' in \emph{In Proceedings of IEEE International
  Conference on Signal Processing, Communications and Computing (ICSPCC),},
  2011.

\bibitem{ImageRev}
J.~B. Allen and D.~A. Berkley, ``Image method for efficiently simulating
  small-room acoustics,'' \emph{Journal of Acoustic Society of America},
  vol.~65, 1979.

\bibitem{MCS}
R.~G. Baraniuk, V.~Cevher, M.~F. Duarte, and C.~Hegde, ``Model-based
  compressive sensing,'' \emph{{IEEE} Transactions in Information Theory},
  2010.

\bibitem{TroppWright}
J.~A. Tropp and S.~J. Wright, ``Computational methods for sparse solution of
  linear inverse problems,,'' \emph{Proceedings of the {IEEE}}, 98, 2010.

\bibitem{ThomasMike}
T.~Blumensath and M.~E. Davies, ``Gradient pursuits,'' \emph{{IEEE}
  Transactions on Signal Processing}, vol.~56, pp. 2370--2382, 2008.

\bibitem{V_CAMSAP}
A.~Kyrillidis and V.~Cevher, ``Recipes on hard thresholding methods,'' in
  \emph{Proceedings of {CAMSAP}}, 2011.

\bibitem{Nestrov10}
Y.~Nesterov, ``A method of solving a convex programming problem with
  convergence rate $\mathcal{O}(1/k^2)$, in soviet mathematics doklady,''
  vol.~27, 1983.

\bibitem{Trop07}
J.~A. Tropp and A.~C. Gilbert, ``Signal recovery from random measurements via
  orthogonal matching pursuit,'' vol. 53(12), 2007.

\bibitem{HMP03}
R.~Gribonval and E.~Bacry, ``Harmonic decomposition of audio signals with
  matching pursuit,'' \emph{{IEEE} Transactions on Signal Processing}, vol.~51,
  pp. 101--111, 2003.

\bibitem{spg_siam}
E.~van~den Berg and M.~P. Friedlander, ``Probing the pareto frontier for basis
  pursuit solutions,,'' \emph{{SIAM} Journal on Scientific Computing}, 2, 2008,
  code available online, \url{http://www.cs.ubc.ca/labs/scl/spgl1}.

\bibitem{hearRoom}
I.~Dokmanic, Y.~Lu, and M.~Vetterli, ``Can one hear the shape of a room: The
  2{-D} polygonal case,'' in \emph{Proceedings of {ICASSP}}, 2011.

\bibitem{CR}
G.~Xu, H.~Liu, L.~Tong, and T.~Kailath, ``A least-squares approach to blind
  channel identification,'' \emph{{IEEE} Transactions on Signal Processing},
  1995.

\bibitem{L1_room}
D.~Ba, F.~Ribeiro, C.~Zhang, and D.~Florencio, ``L1 regularized room modeling
  with compact microphone arrays,'' in \emph{Proceedings of {ICASSP}}, 2010.

\bibitem{AIR}
``Aachen {I}mpulse {R}esponse {(AIR)} database - version 1.2,'' Institute of
  Communication Systems and Data Processing {(IND)}, RWTH Aachen University,
  2010, \url{http://www.ind.rwth-aachen.de/AIR}.

\bibitem{Combettesand11}
P.~L. Combettesand and J.~C. Pesquet, ``Proximal splitting methods in signal
  processing,'' vol.~49, 2011.

\bibitem{book_habet}
E.~A. Habets, ``Speech dereverberation using statistical reverberation
  models,'' \emph{Speech Dereverberation, Springer}, 2010.

\bibitem{Cook55}
R.~K. Cook, R.~V. Waterhouse, R.~D. Berendt, S.~Edelman, and M.~C. Thompson,
  ``Measurement of correlation coefficients in reverberant sound fields,'' vol.
  27(6), 1955.

\bibitem{PF}
C.~Marro, Y.~Mahieux, and K.~U. Simmer, ``Analysis of noise reduction and
  dereverberation techniques based on microphone arrays with postfiltering,''
  \emph{International Workshop on Acoustic Signal Enhancement}, 6, 1998.

\bibitem{PF_all}
T.~Wolff and M.~Buck, ``A generalized view on microphone array postfilters,''
  \emph{International Workshop on Acoustic Signal Enhancement}, 2010.

\bibitem{McCowan03}
I.~A. McCowan and H.~Bourlard, ``Microphone array post-filter based n noise
  field coherence,'' \emph{{IEEE} Transactions on Audio, Speech, and Language
  Processing}, 11(6), 2003.

\bibitem{Carin1}
L.~Carin, ``On the relationship between compressive sensing and random sensor
  arrays,'' \emph{{IEEE} Antennas and Propagation Magazine}, vol.~51, pp.
  72--81, 2009.

\bibitem{Carin2}
L.~Carin, D.~Liu, and B.~Guo, ``Coherence, compressive sensing and random
  sensor arrays,'' \emph{{IEEE} Antennas and Propagation Magazine}, 2011.

\bibitem{MONC}
``The {M}ultichannel {O}verlapping {N}umbers {C}orpus,'' Idiap resources
  available online:, \url{http://www.cslu.ogi.edu/corpora/monc.pdf}.

\bibitem{CVX}
M.~Grant and S.~Boyd, ``{CVX}: Matlab software for disciplined convex
  programming, version 1.21,'' \url{http://cvxr.com/cvx}.

\bibitem{Habets}
\BIBentryALTinterwordspacing
E.~A.~P. Habets, ``Single- and multi-microphone speech dereverberation using
  spectral enhancement,'' Ph.D. dissertation, Technische Universiteit
  Eindhoven, 2007. [Online]. Available:
  \url{http://alexandria.tue.nl/extra2/200710970.pdf}
\BIBentrySTDinterwordspacing

\bibitem{bss_eval}
E.~Vincent, R.~Gribonval, and C.~Fevotte, ``Performance measurement in blind
  audio source separation (code available at
  http://www.irisa.fr/metiss/sassec07/?show=results),'' \emph{{IEEE}
  transactions on audio, speech, and language processing}, vol.~14, 2006.

\bibitem{persia}
L.~D. Persia, D.~Milone, H.~L. Rufiner, and M.~Yanagida, ``Perceptual
  evaluation of blind source separation for robust speech recognition,''
  \emph{Signal Processing, implementation available at},
  \url{http://www.utdallas.edu/~loizou/speech/software.htm}.

\bibitem{asaei3}
A.~Asaei, M.~J. Taghizadeh, H.~Bourlard, and V.~Cevher, ``Multi-party speech
  recovery exploiting structured sparsity models,'' in \emph{Proceedings of
  {INTERPSEECH}}, 2011.

\bibitem{asaei2}
A.~Asaei, H.~Bourlard, and V.~Cevher, ``Model-based compressive sensing for
  multi-party distant speech recognition,'' in \emph{Proceedings of {ICASSP}},
  2011.

\bibitem{Mccowan_overlapping}
D.~C. Moore and I.~A. Mccowan, ``Microphone array speech recognition:
  Experiments on overlapping speech in meetings,'' in \emph{Proceedings of
  {ICASSP}}, 2003.

\bibitem{HTK}
S.~J. Young, D.~Kershaw, J.~Odell, D.~Ollason, V.~Valtchev, and P.~Woodland,
  ``The htk book version 3.4,'' 2006.

\bibitem{Ellis07}
M.~Athineos and D.~P.~W. Ellis, ``Autoregressive modeling of temporal
  envelopes,'' vol. 55(11), 2007.

\bibitem{Elad12}
T.~Peleg, Y.~C. Eldar, and M.~Elad, ``Exploiting statistical dependencies in
  sparse representations for signal recovery,'' vol. 60(5), 2012.

\bibitem{mohammad12}
M.~Golbabaee and P.~Vandergheynst, ``Compressed sensing of simultaneous
  low-rank and joint-sparse matrices,''
  \url{http://infoscience.epfl.ch/record/181506}.

\end{thebibliography}
}
\end{document}